%% file: acl_latex.tex
\pdfoutput=1

\documentclass[11pt]{article}

\usepackage[final]{acl}

\usepackage{times}
\usepackage{latexsym}

\usepackage[T1]{fontenc}

\usepackage[utf8]{inputenc}

\usepackage{microtype}

\usepackage{inconsolata}

\usepackage{graphicx}
\usepackage{booktabs}
\usepackage{xcolor}
\usepackage{multirow}
\usepackage{CJKutf8}
\usepackage{tcolorbox}
\usepackage{colortbl}
\usepackage{pifont}
\usepackage{booktabs}
\usepackage{placeins}
\usepackage{amsmath}
\usepackage{marvosym}

\title{Exploring the Generalizability of Factual Hallucination Mitigation via Enhancing Precise Knowledge Utilization}

\author{Siyuan Zhang, Yichi Zhang, Yinpeng Dong, Hang Su\textsuperscript{\Letter} \\
       Dept. of Comp. Sci. and Tech., Institute for AI, Tsinghua-Bosch Joint ML Center, \\
    THBI Lab, BNRist Center, Tsinghua University, Beijing 100084, China \\
       \small{\href{mailto:email@domain}{zhang-sy24@mails.tsinghua.edu.cn}, \href{mailto:email@domain}{suhangss@mail.tsinghua.edu.cn}}
}

\begin{document}
\maketitle
\input{sec/0_abstract}
\input{sec/1_introduction}
\input{sec/2_related_work}
\input{sec/3_method}
\input{sec/4_experiments}
\input{sec/5_conclusion_etc}

\bibliography{custom}

\appendix
\input{sec/X_appendix}



\end{document}

%% file: sec/0_abstract.tex
\begin{abstract}
Large Language Models (LLMs) often struggle to align their responses with objective facts, resulting in the issue of \textbf{factual hallucinations}, which can be difficult to detect and mislead users without relevant knowledge. Although post-training techniques have been employed to mitigate the issue, existing methods usually suffer from poor generalization and trade-offs in other different capabilities. In this paper, we propose to address these by directly augmenting LLM's fundamental ability to precisely leverage its knowledge and introduce \textbf{PKUE} (\textbf{P}recise \textbf{K}nowledge \textbf{U}tilization \textbf{E}nhancement), which fine-tunes the model on self-generated responses to precise and simple factual questions through preference optimization. Furthermore, we construct \textbf{FactualBench}, a comprehensive and precise factual QA dataset containing 181k Chinese data spanning 21 domains, to facilitate both evaluation and training. Extensive experiments demonstrate that PKUE significantly improves LLM overall performance, with consistent enhancement across factual tasks of various forms, general tasks beyond factuality, and tasks in different language.
\end{abstract}

%% file: sec/1_introduction.tex
\section{Introduction}
\begin{figure}[t]
  \includegraphics[width=\columnwidth]{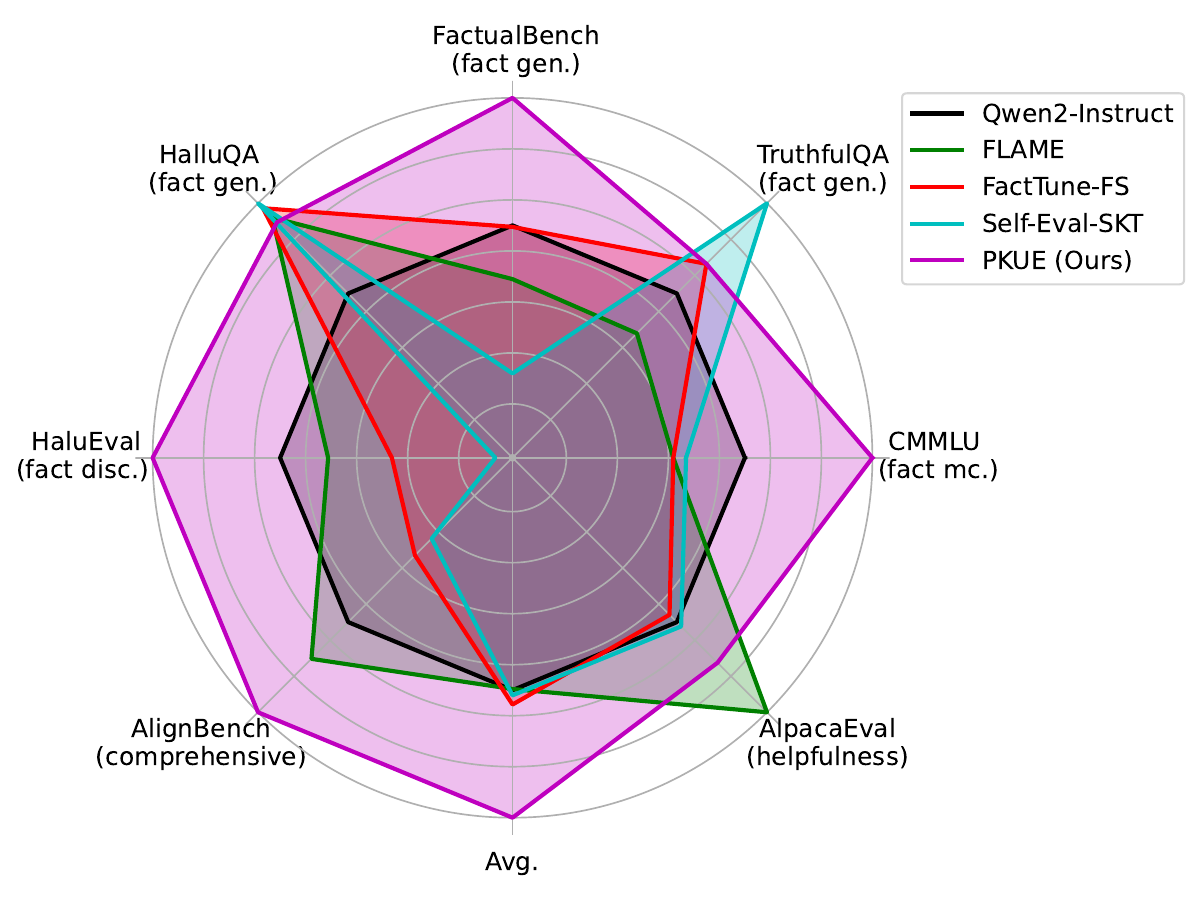}
  \caption{Previous methods on factual hallucination mitigation exhibit poor generalizability across different factual tasks and suffer from degradations on comprehensive abilities and helpfulness, while our method PKUE improves model performance on all seven benchmarks, with a significant advantage on \textit{Avg.}}
  \label{fig:radar}
\end{figure}

\textbf{Factual hallucinations} occur when Large Language Models (LLMs) generate inaccurate or entirely fabricated contents in response to queries \citep{zhang2023siren, huang2023survey}, which can undermine user trust in models and cause significant harm, especially when LLMs are deployed in high-stake applications \citep{ji2023survey, ahmad2023creating, kang2023deficiency}. Furthermore, identifying hallucinations is challenging, as the fabricated contents are often presented plausibly and convincingly, making it difficult for both models and users to recognize inaccuracies \citep{kaddour2023challenges, zhang2023siren}, which emphasizes the essentiality of mitigating hallucinations. 

Among various approaches to mitigate factual hallucinations, from pre-training \citep{gardent2017creating,wang2019revisiting} to inference-time techniques \citep{nakano2021webgpt,dola}, post-training \citep{flame, facttune} has become popular for not requiring large-scale data manipulation or additional runtime computations. Recent methods typically enhance factuality by training on open-ended questions, e.g., ``Tell me a bio of \texttt{an entity}'', which are broad and imprecise. They leave additional space for models to provide answers with diverse contents, subsequently assessed using average factual precision metrics like FActScore \citep{factscore}. However, as shown in Figure~\ref{fig:radar}, these methods lead to declines in other factuality-related tasks and trade-offs in overall performance. The poor generalization can be attributed to the biased signals of these metrics mixing accuracy with length \citep{wei2024long}. Besides, wrong judgments on the correctness of atomic facts, which are caused by the entity ambiguity after response decomposition \citep{chiang2024merging, wanner2024dndscore} and the lack of standard answers in open-ended QA, degrade training effects. Moreover, the inherent noise in the training pairs, as both factually correct and incorrect atomic facts are mixed within a single response, can ultimately reduce the effectiveness of alignment \citep{DBLP:conf/iclr/GuZLL025}. The trade-offs in general performance are not alleviated with additional training on advanced abilities \citep{zhao2023survey} related to factuality under adversarial queries \citep{self-eval-skt} or complex instructions following \citep{flame}, as they are not necessary for other tasks and can lead to forgetting of the acquired abilities \citep{ouyang2022training, lin2024mitigating}.

In this paper, we address the above issues from the perspective of knowledge utilization and propose \textbf{PKUE} (\textbf{P}recise \textbf{K}nowledge \textbf{U}tilization \textbf{E}nhancement) to enhance this capability with the task of precise QA. Since knowledge utilization is a crucial factor for factuality \citep{wang2023survey} and a fundamental ability of LLM \cite{zhao2023survey}, its enhancement is expected to bring generalized improvement beyond hallucination mitigation. We take precise fact-seeking QA as a representative task, which is short-form with standard answers and simple without other attributes besides the correctness. These features make it a proper task to reflect factual hallucination as well as knowledge utilization \citep{roberts2020much, ji2023survey, zhao2023survey} aside from other abilities and prevent the mentioned issues of imprecise open-ended QA. To better enhance utilization rather than inject external information, we leverage the model’s existing knowledge and conduct Direct Preference Optimization (DPO) \citep{rafailov2023direct} training on sampled data from the model itself, which provides more granular bi-directional controls and better generalization \citep{zhang2023siren, chu2025sft} than uni-directional Supervised Fine-tuning (SFT). This self-aligned approach preserves the model’s distribution and limits the post-training shift, avoiding undesirable behaviors \citep{gudibande2023false, zhang2023siren} and additional hallucinations introduced by training on external new knowledge \citep{huang2023survey, flame, gekhman2024does}.

However, a precise factual QA dataset with large scale and diverse domains is lacking for training. Existing ones \citep{yang2015wikiqa, joshi2017triviaqa, yang2018hotpotqa, kwiatkowski2019natural} are usually outdated and fall short in fine-grained domain annotations, limiting their accuracy and diversity. To this end, we build \textbf{FactualBench}, a large-scale dataset with 181k Chinese QA data spanning 21 domains\footnote{We release our dataset in \url{https://github.com/ZSYNOTZSH/FactualBench}}. Chinese is selected since it is a widely used language with a large community but still lacks high-quality datasets. We extract knowledge from the Internet encyclopedia, a widely used pre-training corpus \citep{liu2024datasets, ando2024wikisqe} and can be taken as a knowledge base that LLM has already seen. Multiple filtering strategies are adopted to ensure data quality. Evaluations on FactualBench reveal that while the task is not easy for LLMs, sampling with a higher temperature can yield more correct answers, which leaves space for better utilization of existing knowledge through self-alignment.

Extensive experiments on Qwen2 \citep{qwen2} and Baichuan \citep{yang2023baichuan} show that only PKUE achieves consistent improvement on seven benchmarks covering factuality, helpfulness, and general skills in different forms and languages, presenting the best generalization. Notably, PKUE obtains 4$\times$ and 9$\times$ average improvement compared to existing methods \citep{factscore, flame, self-eval-skt}. More ablation studies confirm the choices of self-generated data and DPO training. Our work proves that improving the knowledge utilization on solely precise and simple QA can promote generalized enhancement, spanning from generative tasks to diverse forms of factual tasks, from factual tasks to other general tasks, and demonstrating cross-lingual transferability from Chinese to English tasks.

%% file: sec/2_related_work.tex
\section{Related Works}
\textbf{Factual hallucination mitigation.} Several studies \citep{wang2019revisiting, gardent2017creating} have explored mitigating hallucinations by improving the quality of pre-training data. But processing vast datasets is time-consuming \citep{zhang2023siren} and is not applicable for models that have completed training. Other approaches \citep{dola, zhang2023alleviating, li2023inference, lee2022factuality} focus on inference-time enhancement, yet these strategies aim for specific tasks and have limited generalization \citep{self-eval-skt}, along with more difficulty generating fluent or diverse texts \citep{ji2023survey}. Furthermore, methods \citep{nakano2021webgpt, gou2024critic} that utilize retrieval-augmented (RAG) techniques introduce significant system complexity \citep{facttune} and depend heavily on the quality of external knowledge bases \citep{zhang2023siren}. Additionally, post-training LLM through SFT \citep{elaraby2023halo,alignmentforhonesty} and Reinforcement Learning \citep{ouyang2022training,kang2024unfamiliar} exhibits a promising reduction in factual error rates. Recently, \citet{facttune, flame, self-eval-skt} use preference learning on self-generated responses. They mainly focus on open-ended questions and rate responses by first adopting external models to split responses into atomic facts, then verifying each fact via RAG \citep{facttune, flame} or a model fine-tuned on millions of related data \citep{self-eval-skt}. This leads to significant complexity, especially when responses contain hundreds of atomic facts. In contrast, PKUE targets precise QA with standard answers, simplifying verification, where no additional training or external databases are required. Moreover, the effects of these methods fail to generalize to other tasks related to factuality and lead to trade-offs in different abilities, while PKUE achieves consistent improvement on them.

\textbf{Precise factual QA tasks} include discriminative, multiple-choice, and generative forms. The former two \citep{thorne2018fever, MMLU, liu2022token, mishra2024fine} have a limited answer space that allows models to guess the correct answer by chance, and therefore are unable to accurately judge whether the corresponding knowledge is possessed. Generative datasets designed with adversarial intents \citep{lin2022truthfulqa, cheng2023evaluating} can effectively provoke hallucinations but tend to focus on specific scenarios, limiting their capacities to reflect performance on more general questions. While large simple generative QA datasets \citep{yang2015wikiqa, joshi2017triviaqa, yang2018hotpotqa, kwiatkowski2019natural} exist, they are mostly built years ago with no domain annotations. In contrast, our annotated dataset offers a comprehensive and up-to-date assessment.

%% file: sec/3_method.tex
\section{Method}
\begin{figure*}[t]
  \includegraphics[width=\linewidth]{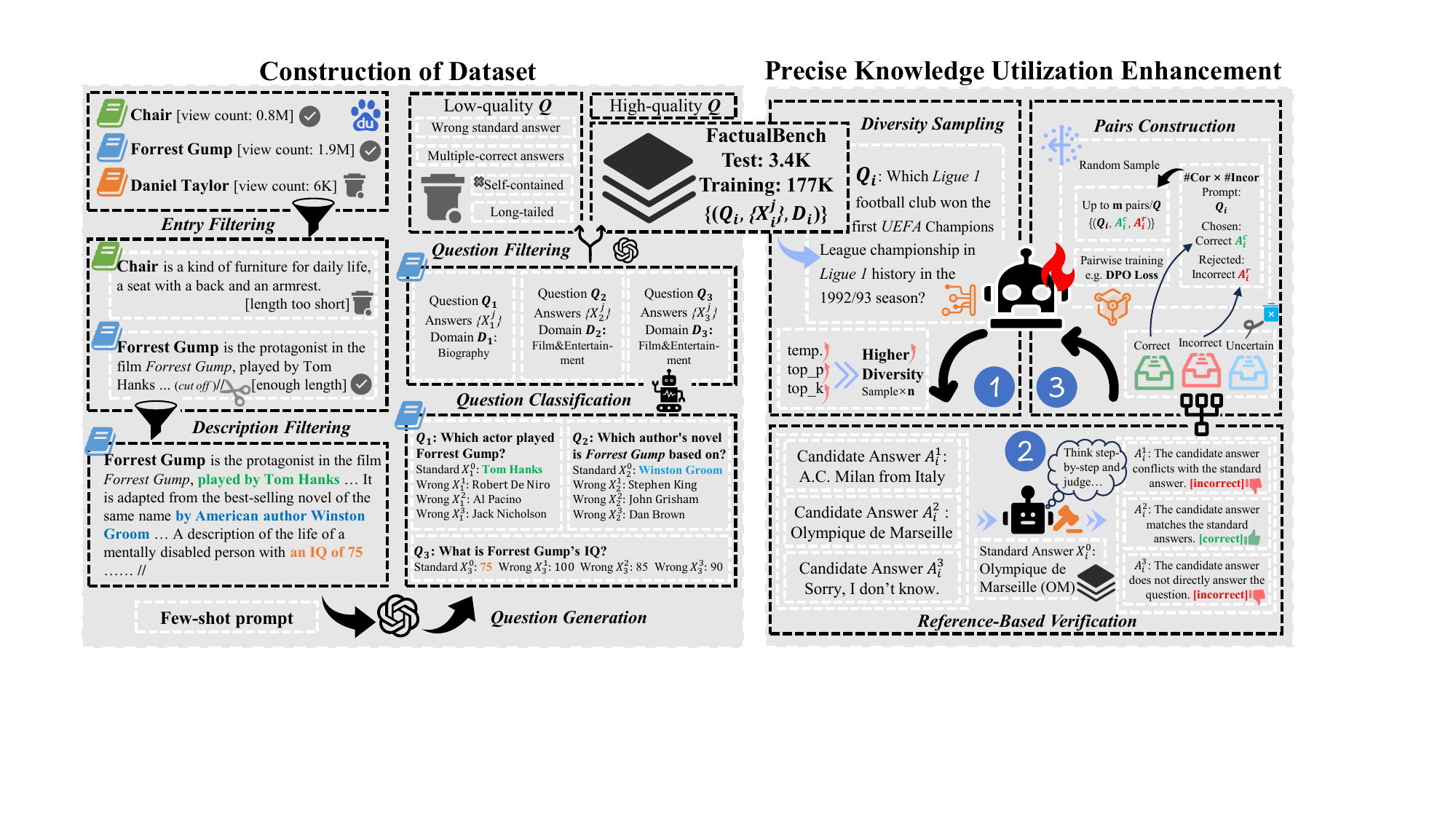}
  \caption{The framework of our work. \textbf{Left}: We first extract factual knowledge from the Internet encyclopedia and construct a large and comprehensive dataset, FactualBench. Several filtering strategies are adopted for higher quality. \textbf{Right}: Next, we align LLM on self-generated response pairs on FactualBench. We elicit diverse responses to the same question, verify each correctness comparing to the standard answer, and sample preference pairs for training.}
  \label{fig:main}
\end{figure*}

To mitigate factual hallucinations and prevent trade-offs in other abilities beyond factuality, we propose PKUE to augment the model's utilization of its existing knowledge. For training and evaluation, we build FactualBench consisting of precise and simple QA data without malicious or misleading adversarial intents. In this section, we will introduce the dataset and the alignment method in detail.

\subsection{FactualBench}
The Internet encyclopedia is selected as the source of the knowledge base since it contains various factual information across domains \citep{wang2023survey, bai2024coig}, which is also a commonly used corpus in LLM pre-training \citep{liu2024datasets, ando2024wikisqe}. Specifically, we use Baidu baike\footnote{\url{https://baike.baidu.com/}}, a prominent encyclopedia in Chinese community, and design a model-based pipeline to generate a large volume of data efficiently, adopting GPT4\footnote{We use the version of gpt-4-0125-preview in this paper.} \citep{achiam2023gpt} and Baichuan model\footnote{\url{https://www.baichuan-ai.com/}} for their strong instruction following capabilities.

During pre-construction experiment, we observe four typical types of low-quality data. 1) Long-tailed questions with obscure and useless knowledge. 2) Questions with multiple correct answers. This is primarily due to imprecise terms in questions that invite subjective judgments and the existence of more valid answers beyond encyclopedia knowledge. 3) Questions with incorrect standard answers. The model may extract knowledge falsely, which is frequent when paragraphs are extremely long or difficult to understand. Some questions fall into this category because they are time-sensitive, but the knowledge in the encyclopedia is outdated. 4) Questions that are not self-contained. Questions containing vague pronouns or ambiguous nouns with multiple interpretations, e.g., abbreviations without clear contexts, will confuse answerers. To guarantee data quality, we then apply few-shot prompts to guide the model and adopt several filtering strategies to remove the low-quality data.

\textbf{Construction and Composition.} As illustrated in Figure \ref{fig:main} (left), FactualBench is constructed in five steps. \textit{1) Entry filtering}. We initially sample millions of entries from the publicly available encyclopedia, ensuring broad coverage over subjects and domains. For each entry, we retain its object, view count, and brief description. To avoid generating questions on long-tailed knowledge, we set a view count threshold of 0.5M, and 89,658 entries remain after this filtering. \textit{2) Description filtering}. The performance of the model tends to decrease as the context length increases \citep{liu2024lost, sun2023survey, li2024long}. Excessively lengthy descriptions can provide superfluous information and lead to low-quality responses. Conversely, overly brief descriptions lack sufficient factual information. To balance this, we filter out descriptions shorter than 100 characters and truncate those exceeding 800 characters. 64,315 entries remain after this process. \textit{3) Question generation}. We instruct GPT4 to generate up to three precise questions per truncated description. For each question $Q_i$, GPT4 is also required to provide one standard answer $X_i^0$ and three wrong answers $\{X_i^j\}_{j=1}^3$ for further evaluation and training uses. To ensure adherence to our instructions, we add two examples for few-shot prompting. A total of 192,927 QA samples are generated in this process. \textit{4) Question classification}. A domain classifier based on Baichuan model, fine-tuned on massive high-quality data, is employed to categorize all generated questions into different domains $D_i$. We maintain domains containing more than 500 questions and uniformly categorize the rest as \emph{others}. We explain how we obtain the classifier in Appendix \ref{cha:classifier obtaining}. \textit{5) Question filtering}. We query GPT4 once again to filter out low-quality questions. Each question is assessed independently without the corresponding description, and GPT4 is instructed to identify whether the question falls into one of the low-quality types mentioned above through step-by-step reasoning. 

\begin{CJK*}{UTF8}{gbsn}
\begin{table}[!tb]
  \centering
    \resizebox{\columnwidth}{!}{%
    \begin{tabular}{cc}
    \toprule[1.5pt]
    \multirow{2}{*}{Question $Q_i$} & 第一台微波量子放大器是在哪一年制成的？\\
    & In which year was the first microwave quantum amplifier made? \\ 
    \midrule
    \multirow{2}{*}{Standard Answer $X_i^0$} & 第一台微波量子放大器是在1954年制成的。\\
    & The first microwave quantum amplifier was made in 1954. \\
    \midrule
    \multirow{2}{*}{Wrong Answer $X_i^1$} & 第一台微波量子放大器是在1958年制成的。\\
    & The first microwave quantum amplifier was made in 1958. \\
    \multirow{2}{*}{Wrong Answer $X_i^2$} & 第一台微波量子放大器是在1960年制成的。\\
    & The first microwave quantum amplifier was made in 1960. \\
    \multirow{2}{*}{Wrong Answer $X_i^3$} & 第一台微波量子放大器是在1962年制成的。\\
    & The first microwave quantum amplifier was made in 1962. \\ \midrule
    \multirow{2}{*}{Domain $D_i$} & 高新科技 \\
    & high technology \\
    \bottomrule[1.5pt]
    \end{tabular}%
    }
  \caption{Each sample in FactualBench contains a question $Q_i$, a standard answer $X_i^0$, 3 wrong answers $\{X_i^j\}$ and a domain $D_i$ it belongs to. The English translation is for reference. Appendix \ref{cha:more examples} presents more examples.}
  \label{tab:example}
\end{table}
\end{CJK*}

Finally, 181,176 questions are reserved, where assessments of 1,000 samples indicate that an approximately 86\% high-quality rate is acquired. To evaluate the LLMs' ability to utilize knowledge, we randomly select a subset of questions for the test set. We do selection taking each entry (entries containing \textit{others} domain questions are excluded) as a unit to maintain that all questions in the test set are separate from the training set, and restrict each domain to a similar number of questions. 3,462 questions are selected, and the remaining 177,714 samples form the training set. We manually refine low-quality questions in the test set after selection to ensure its high quality. Specifically, we provide annotators with the QA pairs and entry references, requiring them to determine whether a QA pair falls into one of the low-quality cases and rewrite the question or the answer if needed. We present the construction prompts in Appendix \ref{cha:generation prompts}, a sample in Table \ref{tab:example}, and the domain distribution in Table \ref{tab:construction}.

\begin{CJK*}{UTF8}{gbsn}
\begin{table}[!tb]
  \centering
   \resizebox{\columnwidth}{!}{%
    \begin{tabular}{ccccc}
    \toprule[1.5pt]
    Domain    & 中文名   & Test & Training & Total \\
    \midrule
    film\&entertainment     & 影视娱乐  &   201        &   54,489       &  54,690     \\
    education\&training           & 教育培养  &   161        &   3,703       &  3,864     \\
    physics, chemistry, mathematics\&biology   & 数理化生  &    201 &   9,189    &  9,390     \\
    history\&traditional culture    & 历史国学  &    202       &   18,108       &  18,310     \\
    biography           & 人物百科  &    201       &   11,844       &  12,045     \\
    politics\&law            & 政治法律  &   175        &   6,368       &  6,453     \\
    economics\&management         & 经济管理  &    160       &   4,543       &  4,703     \\
    computer science           & 计算机科学 &  201         &  6,253        &  6,454     \\
    medical                   & 医学    &     167      &   7,073       &   7,240    \\
    sociology\&humanity        & 社会人文  &    199       &    8,503      &  8,702     \\
    agriculture, forestry, fisheries\&allied industries & 农林牧渔  & 153 & 3,728 &  3,881  \\
    astronomy\&geography          & 天文地理  &    160       &   3,896       &  4,056     \\
    sports\&tourism            & 运动旅游  &    157       &   4,869       &  5,026     \\
    digital\&automotive         & 数码汽车  &    176       &   3,887       &  4,063     \\
    industrial engineering       & 工业工程  &   172        &   3,283       &  3,455     \\
    military\&war                 & 军武战争  &    151       &   2,569       &  2,720     \\
    slang\&memes                    & 网词网梗  &    151       &   529       &  680     \\
    work\&life                      & 工作生活  &    174       &   5,853       &  6,027     \\
    high technology                  & 高新科技  &    150       &   310       &  460     \\
    religion\&culture           & 信仰文化  &     150      &   510       &  660     \\
    others                   & 其他    &    -       &   18,207       &  18,207     \\ 
    \midrule
    total                     & -     &   3,462        &   177,714       &   181,176   \\
    \bottomrule[1.5pt]
    \end{tabular}%
    }
  \caption{Domain distribution of FactualBench.}
  \label{tab:construction}
\end{table}
\end{CJK*}

\textbf{Evaluation.} Following previous works \citep{liu2023alignbench, zheng2023judging}, a model-based approach is employed to expedite the evaluation. Note that rule-based automatic metrics such as ROUGE \citep{lin2004rouge} and BLEU \citep{papineni2002bleu} have been shown to exhibit significant biases in assessment \citep{lou2024large}, we judge the correctness of the answer at semantic-level. The verifier is supposed to focus solely on the content directly addressing the question and ignore the extraneous information. A response is considered correct only when it indeed answers the question (rather than ``I don't know'') and matches the standard answer. This is reasonable since the model is expected to have been trained on the relevant, frequently viewed data and should possess the necessary knowledge, and the portion of evasive answers only counts for approximately 1\%, which affects the evaluation result lightly. To improve judgment accuracy, we provide several examples and instruct the verifier to offer analysis before making the final decision. GPT4 is chosen as the verifier, which achieves a 96\% consistency with humans, validating the effectiveness. Furthermore, we perform five independent evaluations on the same generated answers, observing a deviation of only 0.4\% between the highest and lowest accuracies, thereby confirming the robustness and stability of our evaluation procedure. We present the evaluation prompt in Appendix \ref{cha:evaluation prompts}.

14 popular LLMs are evaluated on FactualBench: Baichuan series \citep{yang2023baichuan}, Qwen series \citep{qwen, qwen2}, Llama-3 series \citep{llama3modelcard}, Yi \citep{ai2024yi}, Command-R series \citep{CommandR, CommandR+}, DeepSeek \citep{deepseekv2}, and GPT4, where we prioritize the chat / instruct versions. We list the brief results in Table \ref{tab:eval result}. The accuracy (Acc.) on our test set ranges from 39.11\% to 75.62\%, indicating that LLMs still have deficiencies in the basic factual QA task. Detailed domain-level accuracy and additional analyses of the results can be found in Appendix \ref{cha:detailed result}. 

\begin{table}[t]
  \centering
    \resizebox{\columnwidth}{!}{%
    
    \begin{tabular}{cc}
    \toprule[1.5pt]
     Model & Acc. \\ \midrule
     \bf Baichuan1 & 48.24 \\  
     \bf Baichuan2 & 55.37 \\ 
     \bf Qwen1.5-7B & 48.87 \\ 
     \bf Qwen2-7B & 56.27 \\  
     Llama-3-8B & 39.11 \\ 
     \bottomrule[1.5pt]
   \end{tabular}%
   
   \begin{tabular}{|cc|}
    \toprule[1.5pt]
     Model & Acc. \\ \midrule
     \bf Baichuan3 & 67.50 \\
     \bf Yi-34B & 67.30 \\
     Command-R 35B & 54.30 \\
     Llama-3-70B & 49.65 \\
     \bf Qwen2-72B & 73.71 \\
     \bottomrule[1.5pt]
   \end{tabular}%
   
   \begin{tabular}{cc}
    \toprule[1.5pt]
     Model & Acc. \\ \midrule
     \bf Baichuan4 & 75.07 \\
     Command-R+ 104B &  60.17 \\
     \bf DeepSeek-v2 & \multirow{2}{*}{75.62} \\
     \bf -0628 MoE-236B & \\
     GPT4 & 65.71 \\
     \bottomrule[1.5pt]
   \end{tabular}%
   
    }
  \caption{Performance on FactualBench rated by GPT4. Models in bold are proficient in Chinese.}
  \label{tab:eval result}
\end{table}

\subsection{PKUE}
\begin{figure}[ht]
  \includegraphics[width=\columnwidth]{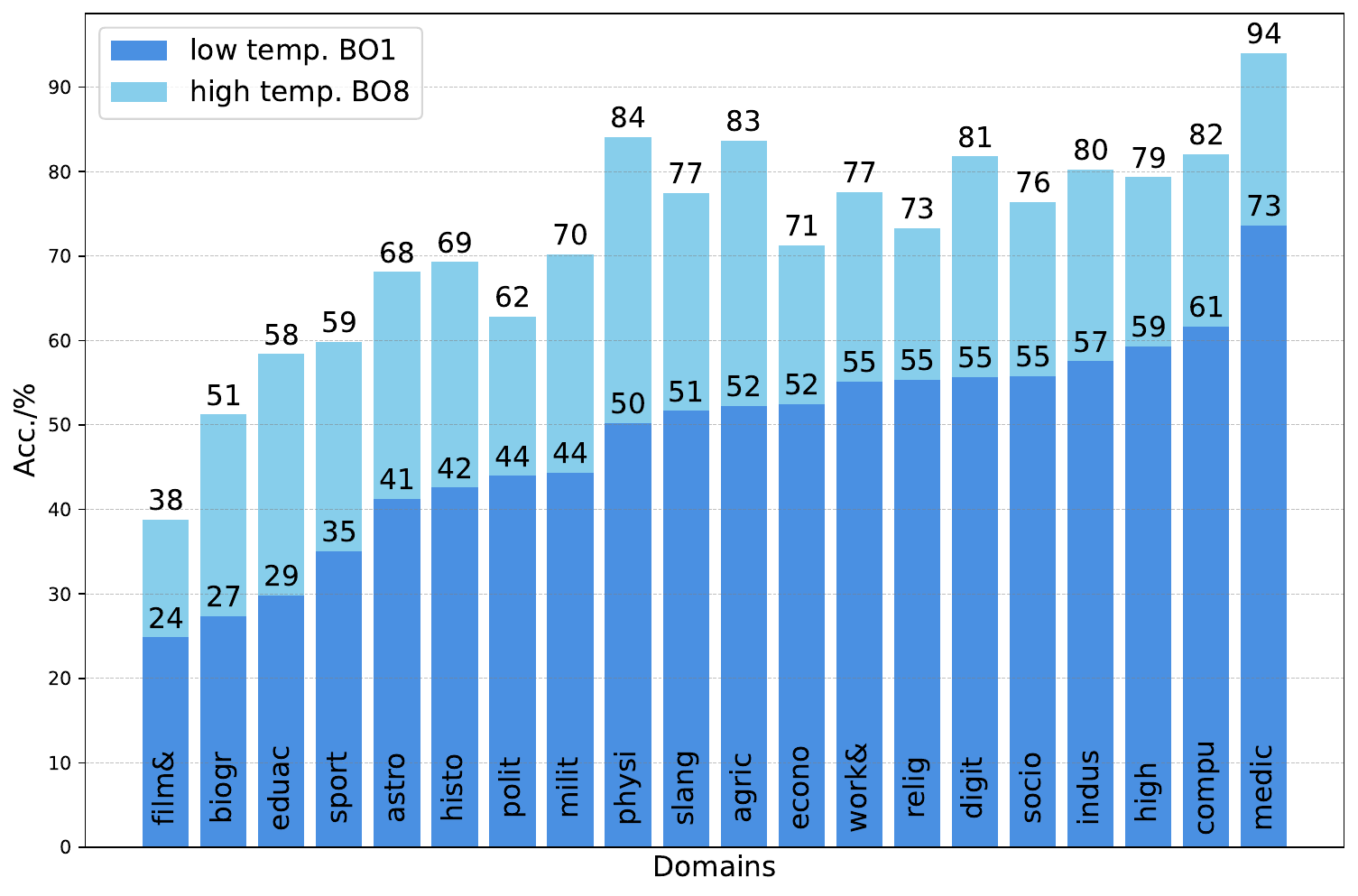}
  \caption{A comparison between Baichuan1 accuracy in \textit{low temp. BO1} and \textit{high temp. BO8}. Significant gaps in all domains demonstrate the potential of the model. Each domain is represented by its first five letters.}
  \label{fig:BO1andBO8}
\end{figure}

For cases where the LLM initially provides incorrect responses, we observe that it can generate correct answers when given greater output diversity. Taking Baichuan1 as an example, we increase the response variability by increasing the generation temperature and sampling the model's responses eight times (\textit{high temp. BO8}, where \textit{BO} stands for \textit{Best of}), contrasting with the standard inference setting (\textit{low temp. BO1}). Given the extensive answer space in the generative task, it is statistically improbable for a model to randomly guess the correct answer, so we consider the model to possess relevant knowledge if at least one of the generated responses is correct. As illustrated in Figure \ref{fig:BO1andBO8}, comparison between \textit{BO8} and \textit{BO1} reveals a substantial portion of the model's capabilities remains underutilized, indicating an untapped potential in the utilization of knowledge. This also verifies the feasibility of building pairs on self-generated responses. Some cases are provided in Appendix \ref{cha:BO8 cases}.

To stimulate the potential and enhance the precise knowledge utilization of models, we propose PKUE that aligns models on self-generated responses to precise and simple QA through preference learning. As shown in Figure \ref{fig:main} (right), the alignment includes three phases. \textit{1) Diversity Sampling.} For each question $Q_i$ in FactualBench training set ${\mathcal{D}^{\text{train}}}$, we sample $n$ responses from the model $\pi$ in higher diversity by increasing generation configurations such as temperature, top-p, and top-k. \textit{2) Reference-Based Verification.} The collected candidate responses are then provided to a verifier model, together with the standard answer $X_i^0$ from FactualBench. The verifier evaluates the responses after carefully analyzing, which acts as a judge function $\mathcal{J}$ to output $1$ or $0$ indicating correctness or not. Each evaluation result is formatted in a consistent manner to facilitate subsequent classification. \textit{3) Pairs Construction.} We classify all responses according to their correctness, discarding those with uncertain evaluation results (due to verifier failing in instruction following or low quality of questions), and construct a set as follows: 
\begin{equation}\textstyle
    \{(\text{Prompt } Q_i, \text{Chosen } A_i^c, \text{Rejected } A_i^r)\},
\end{equation}
which is under the following constraint conditions:
\begin{align}\textstyle
    (Q_i, X_i^0) \sim {\mathcal{D}^{\text{train}}}; A_i^c, A_i^r \sim \pi (\cdot | Q_i);\\
    \mathcal{J}(Q_i, A_i^c, X_i^0) = 1;  \mathcal{J}(Q_i, A_i^r, X_i^0) = 0.
\end{align}
However, different questions can contribute significantly varying numbers of preference pairs ($=$ correct count $\times$ incorrect count). To balance this disparity, we randomly down-sample up to $m$ pairs for each question, which compose the tuning set.

In this way, we can quickly generate a tuning set $\mathcal{D}^{\text{tuning}}$ containing massive data without human intervention. Then we fine-tune the model on the tuning data through preference learning, DPO \citep{rafailov2023direct}, whose loss is defined as follows:
\begin{equation}\textstyle
    -E \left[ \log \sigma ( \beta \log \frac{\pi_{\theta} (A_i^c | Q_i)} {\pi_{\text{ref}} (A_i^c | Q_i)} - \beta \log \frac {\pi_{\theta} (A_i^r | Q_i)} {\pi_{\text{ref}} (A_i^r | Q_i)}) \right],
\end{equation}
where $(Q_i, A_i^c, A_i^r) \sim \mathcal{D}^{\text{tuning}}$, $\pi_{\theta}$ is the optimal model initialized in model $\pi$ before optimization, while $\pi_{\text{ref}}$ is the frozen $\pi$. $\sigma$ denotes the sigmoid function, and $\beta$ is a hyperparameter.

%% file: sec/4_experiments.tex
\section{Experiments}
In this section, we present the training results using PKUE. Comparison with the other three baselines validates our effectiveness, and more ablation studies are conducted to investigate how our detailed settings influence training outcomes.

\subsection{Settings}
We use Qwen2-7B-Instruct \citep{qwen2} and Baichuan1-Chat as experimental base models. To have a comparable training computation with baselines, we randomly sample a small split from the FactualBench training set, containing 24k samples, which we denote as \textit{(small)}. Since verification can be costly and time-consuming frequently visiting GPT4 through API, we adopt weaker models, Qwen and Baichuan, as verifiers, respectively, to accelerate the process. These models still have acceptable judgment accuracy since standard answers are also provided. We have a discussion on verifier accuracy in Appendix \ref{cha:verifier accuracy}. For each question, we sample $n=8$ responses from the model and reserve up to $m=8$ preference pairs for the tuning set. We set top-k$=$50, top-p$=$0.9, temperature$=$1.4 for Qwen2, and temperature$=$1.2 for Baichuan1. Training details are provided in Appendix \ref{sec:training details}.

For baselines, we select FLAME \citep{flame}, FactTune-FS \citep{facttune}, and Self-Eval-SKT \citep{self-eval-skt}, all of which aim to enhance factuality. These methods involve training on open-ended questions and additional attention on instruct-following queries \citep{kopf2023openassistant} in \citet{flame} or adversarial questions \citep{lin2022truthfulqa} in \citet{self-eval-skt}. FLAME adopts 23,200 prompts for training, and for the remaining two baseline methods, we scale up their training prompts to ensure comparability with our approach. We reproduce their training procedures on Qwen2 and Baichuan1 adhering to the settings in their original papers.

We adopt FactualBench to evaluate factuality on precise and simple QA, with more benchmarks assessing factuality across different tasks: TruthfulQA \citep{lin2022truthfulqa} and HalluQA \citep{cheng2023evaluating} for generative tasks (gen.) and factuality to adversarial questions, CMMLU \citep{li2023cmmlu} for multiple-choice task (mc.), and HaluEval \citep{li2023halueval} for discriminative task (disc.). Additionally, we adopt AlignBench \citep{liu2023alignbench} containing eight sub-tasks for comprehensive advanced abilities and AlpacaEval \citep{alpaca_eval} for helpfulness to reflect the broader impact of training beyond factuality. We report the average score (out of 10) for AlignBench, the win rate (\%) against the base model for AlpacaEval, and the accuracy (\%) for the remaining ones. Since Self-Eval-SKT uses partial data from TruthfulQA, we report the accuracy on the rest of the data for this method. We calculate \textit{Avg.} averaging performance on the benchmarks, where the AlignBench score is multiplied by 10 to align with other metrics, and AlpacaEval is excluded due to its relative metric. More details on the evaluation are provided in Appendix \ref{cha:benchmark detail}.

\subsection{Main Results}
\begin{table*}[ht]
  \centering
   \resizebox{\textwidth}{!}{%
    \begin{tabular}{l|ccccc|ccccccccc|c|c}
    \toprule[1.5pt]
     Model & \rotatebox{80}{\textbf{FactualBench} (gen.)} & \rotatebox{80}{\textbf{TruthfulQA} (gen.)} & \rotatebox{80}{\textbf{HalluQA} (gen.)} & \rotatebox{80}{\textbf{CMMLU} (mc.)} & \rotatebox{80}{\textbf{HaluEval} (disc.)}  & \rotatebox{80}{\textbf{AlignBench}} & \rotatebox{80}{--Prof. Knowledge} & \rotatebox{80}{--Mathematics} & \rotatebox{80}{--Fundamental Lang.} & \rotatebox{80}{--Logical Reasoning} & \rotatebox{80}{--Understanding} & \rotatebox{80}{--Writing} & \rotatebox{80}{--Role Play} & \rotatebox{80}{--Open-ended} & \rotatebox{80}{AlpacaEval (helpful)} & \rotatebox{80}{\textbf{$\Delta$Avg.}} \\ 
     \midrule
     \multicolumn{17}{c}{\sc Qwen2-7B-Instruct} \\ 
    \midrule
     Base  & 56.27 & 52.75 & 46.44 & 80.85 & 52.30 & 6.69 & 6.62 & 6.65 & 6.51 & 5.07 & 6.76 & 7.15 & 7.59 & 7.46 & 50.00 & -  \\ 
     \midrule
     FLAME  & \textcolor{red}{55.20} & \textcolor{red}{50.43} & 50.00 & \textcolor{red}{80.12} & \textcolor{red}{51.66} & 6.80 & \textcolor{red}{6.59} & \textcolor{red}{6.22} & 6.60 & \bf 5.83 & 6.78 & 7.31 & \bf 7.85 & 7.72 & \bf 68.32 & \textcolor{red}{-0.02}  \\
     FactTune-FS  & \textcolor{red}{56.24} & 54.47 & 50.44 & \textcolor{red}{80.12} & \textcolor{red}{50.81} & \textcolor{red}{6.49} & \textcolor{red}{6.35} & \textcolor{red}{6.37} & \textcolor{red}{6.32} & 5.14 & \textcolor{red}{6.31} & \textcolor{red}{6.77} & \textcolor{red}{7.49} & \textcolor{red}{7.45} & \textcolor{red}{48.51} & +0.24  \\
     Self-Eval-SKT & \textcolor{red}{53.32} & \bf 57.99 & \bf 50.67 & \textcolor{red}{80.25} & \textcolor{red}{49.43} & \textcolor{red}{6.44} & \textcolor{red}{6.40} & 6.67 & \textcolor{red}{6.27} & 5.08 & \textcolor{red}{6.17} & \textcolor{red}{6.84} & \textcolor{red}{7.09} & \textcolor{red}{7.10} & 50.87 & +0.09 \\ 
     \midrule
     PKUE (small) & \bf 58.81 & 54.47 & 49.78 & \bf 82.15 & \bf 54.00 & \bf 6.96 & \bf 6.63 & \bf 6.94 & \bf 6.94 & 5.56 & \bf 6.93 & \bf 7.43 & 7.84 & \bf 7.92 & 58.26 & \bf +2.22  \\ 
     \midrule
    \multicolumn{17}{c}{\sc Baichuan1-Chat} \\ 
    \midrule
     Base  & 48.24 & 30.23 & 32.00 & 48.85 & 50.35 & 5.03 & 5.34 & 2.71 & 5.57 & 3.20 & 5.86 & 6.32 & 6.33 & 6.63 & 50.00 & -  \\ 
     \midrule
     FLAME  & 51.16 & \textcolor{red}{29.62} & 32.00 & 49.33 & \bf 51.28 & 5.21 & 5.80 & 2.85 & 5.65 & \bf 3.43 & \bf 6.05 & \textcolor{red}{6.21} & 6.38 & \bf 7.00 & \bf 56.46 & +0.92  \\
     FactTune-FS  & 50.43 & 31.95 & \textcolor{red}{30.89} & 48.94 & 50.93 & \textcolor{red}{4.29} & \textcolor{red}{4.56} & \textcolor{red}{2.17} & \textcolor{red}{4.12} & \textcolor{red}{2.51} & \textcolor{red}{4.98} & \textcolor{red}{5.45} & \textcolor{red}{5.76} & \textcolor{red}{6.37}  & 52.24 & \textcolor{red}{-0.66}  \\
     Self-Eval-SKT & 48.41 & \bf 36.11 & 33.33 & 49.24 & \textcolor{red}{50.29} & \textcolor{red}{4.83} & 5.37 & 2.76 & \textcolor{red}{5.09} & 3.39 & \textcolor{red}{5.57} & \textcolor{red}{5.75} & \textcolor{red}{5.84} & \textcolor{red}{6.11}  & 54.84 & +0.95 \\ 
     \midrule
     PKUE (small) & \bf 57.37 & 33.78 & \bf 38.44 & \bf 50.13 & 50.63 & \bf 5.30 & \bf 5.92 & \bf 3.02 & \bf 5.66 & 3.37 & 5.97 & \bf 6.53 & \bf 6.55 & 6.79 & 54.84 & \bf +3.90  \\ 
     \midrule
     PKUE (full) & \underline{58.29} & 35.86 & \underline{38.89} & \underline{50.92} & \underline{52.05} & \underline{5.38} & \underline{6.25} & \underline{3.03} & \underline{5.76} & \underline{3.55} & \underline{6.12} & 6.52 & 6.36 & 6.79 & \underline{63.99} & \underline{+4.97}  \\ 
     \bottomrule[1.5pt]
    \end{tabular}
    }
  \caption{\label{tab:main result}
    Performance on benchmarks reflecting factuality, helpfulness, and comprehensive abilities. We mark the decreased results in \textcolor{red}{red}, and the best results except PKUE (full) in \textbf{bold} (if PKUE (full) achieves even better result, it is marked in \underline{underline}). Sub-tasks of AlignBench are listed in abbreviation. Domain-level accuracy on FactualBench is shown in Appendix \ref{cha:detailed main experiment result}
  }
\end{table*}

Table \ref{tab:main result} presents the performance of different methods. All baselines have decreased performance not only on factuality-related tasks but also on advanced skills and helpfulness, highlighting the deficiency in generalization. In contrast, PKUE leads to consistent improvement across all benchmarks, including all sub-tasks in AlignBench. Specifically, PKUE achieves 2.22 and 3.90 improvement in $Avg.$ on Qwen2 and Baichuan1, respectively, $4 \times$ and $9 \times$ to the best baselines. PKUE also achieves the best results on almost all benchmarks, except TruthfulQA, HalluQA, and AlpacaEval, where Self-Eval-SKT and FLAME incorporate in-domain data for training. We also include the results of PKUE on the full FactualBench training set \textit{(full)} to better exploit our dataset, achieving much better results. 

Notably, changes on FactualBench reveal that PKUE stimulates partial potential in the model, while baselines show limited improvement and even declines, which indicates that training on imprecise open-ended questions with average precision metrics offers limited gains in the model's utilization of precise factual knowledge. To better understand how PKUE improves precise knowledge utilization ability, a further experiment on Qwen2-7B is conducted. We randomly sample 500 questions from FactualBench test split and, for each question, prompt the model to generate eight different responses under \textit{high temp.} condition both before and after PKUE training. We categorize these questions based on their response accuracy before training and calculate the average accuracy within each group after training. As shown in Table \ref{tab:further analysis}, questions that can be initially answered correctly exhibit greater stability and higher accuracy after training. For questions that are initially answered totally incorrectly, PKUE training enhances the model's ability to produce the right responses. These results suggest that PKUE not only reinforces the utilization of correct knowledge, but also helps the model discover the appropriate pathway to the correct answer. 

\begin{table}[!t]
    \centering
   \resizebox{\linewidth}{!}{%
    \begin{tabular}{c|ccccccccc}
    \toprule[1.5pt]
    Acc. before training & 0/8 & 1/8 & 2/8 & 3/8 & 4/8 & 5/8 & 6/8 & 7/8 & 8/8 \\
    \midrule
    prompt number & 160 & 38 & 31 & 25 & 30 & 21 & 29 & 39 & 127 \\
    \midrule
    Acc. after training /\% & 3.44 & 17.84 & 27.02 & 40.00 & 55.42 & 70.83 & 77.16 & 89.10 & 98.03 \\
     \bottomrule[1.5pt]
    \end{tabular}
    }
  \caption{\label{tab:further analysis}
    Qwen2-7B accuracy on 500 FactualBench questions before and after training under \textit{high temp.} condition.
  }
\end{table}

A particularly noteworthy observation is that training the model exclusively on the generative simple and precise QA data can lead to broad generalized improvement: 1) Enhanced performance on diverse formats of factuality, including multiple-choice tasks in CMMLU, discriminative tasks in HaluEval, and even adversarial tasks in TruthfulQA, HalluQA. 2) Gains in general capabilities beyond factuality, such as helpfulness in AlpacaEval and comprehensive skills in AlignBench. 3) Cross-lingual generalization from Chinese to English tasks like TruthfulQA and AlpacaEval. These findings underscore the fundamental importance of precise knowledge utilization for various capabilities, and suggest the similarity of abilities across different languages. While the presented results are sufficient to validate the generalizability advantage of PKUE, we additionally conduct experiments on two other models, more factual benchmarks, and compare PKUE with more training-free baselines in Appendix \ref{cha: more experiments} to further strengthen the effectiveness of our method.

\subsection{Ablation Studies}
More ablation studies are conducted to further validate the effectiveness of our settings. Detailed and complete results are shown in Appendix \ref{cha:detailed ablation experiment result}.

\textbf{Ablation on data sources.} Our method adopts self-generated responses to align models, denoted as \textit{self}. In addition, we validate more data sources. The standard answers and wrong answers from the dataset generated by GPT4 are denoted as \textit{dataset}. Model responses given the reference descriptions are denoted as \textit{w/ desc.}, which are generally correct since standard answers are contained in descriptions. We also train Qwen on responses generated by \textit{Baichuan}. For SFT, a single correct label is randomly selected per question. Training results are shown in Table \ref{tab:data source ablation}.

Training on self-generated data yields better results for both DPO and SFT. While SFT on ground truth data (\textit{dataset} and \textit{w/ desc.}) improves performance on FactualBench, it leads to sharp declines on other tasks, which can be attributed to learning on responses with extremely different styles, short and concise, from the model itself. For DPO, training on \textit{dataset} or other model's responses can still achieve competitive results. However, it is crucial to have chosen and rejected labels in the same distribution to prevent reward hacking \citep{shekhar2024see}.

\begin{table}[!t]
  \centering
   \resizebox{\linewidth}{!}{%
  \begin{tabular}{c|cc|cccc}
    \toprule[1.5pt]
    Loss & Chosen & Rejected & FactualBench & AlignBench & AlpacaEval & $\Delta$Avg. \\
    \midrule
    \multicolumn{7}{c}{\sc Qwen2-7B-Instruct} \\ 
    \midrule
    SFT & self & - & 55.43 & 6.63 & 44.22 & -0.66 \\
    SFT & Baichuan & - & 49.97 & 4.98 & 15.03 & -13.61 \\
    SFT & dataset & - & 50.38 & 3.56 & 7.20 & -23.22 \\
    \midrule
    DPO & self & self & \bf 58.81 & \bf 6.96 & \bf 58.26 & \bf +2.22 \\
    DPO & Baichuan & Baichuan & 58.17 & 6.71 & 39.19 & +0.45 \\
    DPO & dataset & dataset & 55.75 & 6.50 & 36.06 & -0.65 \\
    \midrule
    \multicolumn{7}{c}{\sc Baichuan1-Chat} \\ 
    \midrule
    SFT & self & - & 51.33 & 5.04 & 37.58 & +1.29 \\
    SFT & w/ desc. & - & 55.63 & 4.47 & 36.96 & -5.69 \\
    SFT & dataset & - & 55.86 & 3.73 & 26.65 & -10.18 \\
    \midrule
    DPO & self & self & \bf 58.29 & \bf 5.38 & \bf 63.99 & \bf +4.97 \\
    DPO & w/ desc. & self & 18.17 & 4.07 & 32.80 & -13.67 \\
    DPO & dataset & self & 5.40 & 3.28 & 19.07 & -21.56 \\
    DPO & dataset & dataset & 49.08 & 4.82 & 39.07 & -1.40 \\
    \bottomrule[1.5pt]
  \end{tabular}
  }
  \caption{Results after training on different data sources.}
  \label{tab:data source ablation}
\end{table}

\textbf{Ablation on loss functions.} We choose DPO for its fine-grained bi-directional signals, and SFT training is conducted for effectiveness comparison. Beyond SFT on a single label per question (\textit{single label}), we also explore SFT the model on all correct answers (\textit{all labels}). Moreover, existing researches suggest that fusing DPO with SFT loss can help mitigate overoptimization on rejected labels \citep{he2024complex, liu2024provably}, which we denote as \textit{SFT+DPO}. Furthermore, additional SFT training prior DPO on the tuning set is supposed to reduce distribution shift issues and thus help training \citep{xu2024dpo}, which we denote as \emph{SFT then DPO}. All training is conducted on self-generated data. Training results are shown in Table \ref{tab:loss ablation}.

The comparison between DPO and SFT shows that preference data will lead to greater improvement, even for \textit{DPO (small)} with fewer tuning data than \textit{SFT (single label)}, confirming that uni-directional signal is indeed insufficient for our task. Additionally, the difference between SFT on \textit{single label} and \textit{all labels} demonstrates that more labels for the same question in SFT will not enhance training effectiveness. Moreover, neither \textit{SFT then DPO} nor \textit{SFT+DPO} outperforms DPO. Since the data are sampled from the model itself, there is little distribution shift and a low likelihood of having reward hacking solely on rejected labels during training, which emphasizes the stability of our method.

\begin{figure*}
  \includegraphics[width=0.24\textwidth]{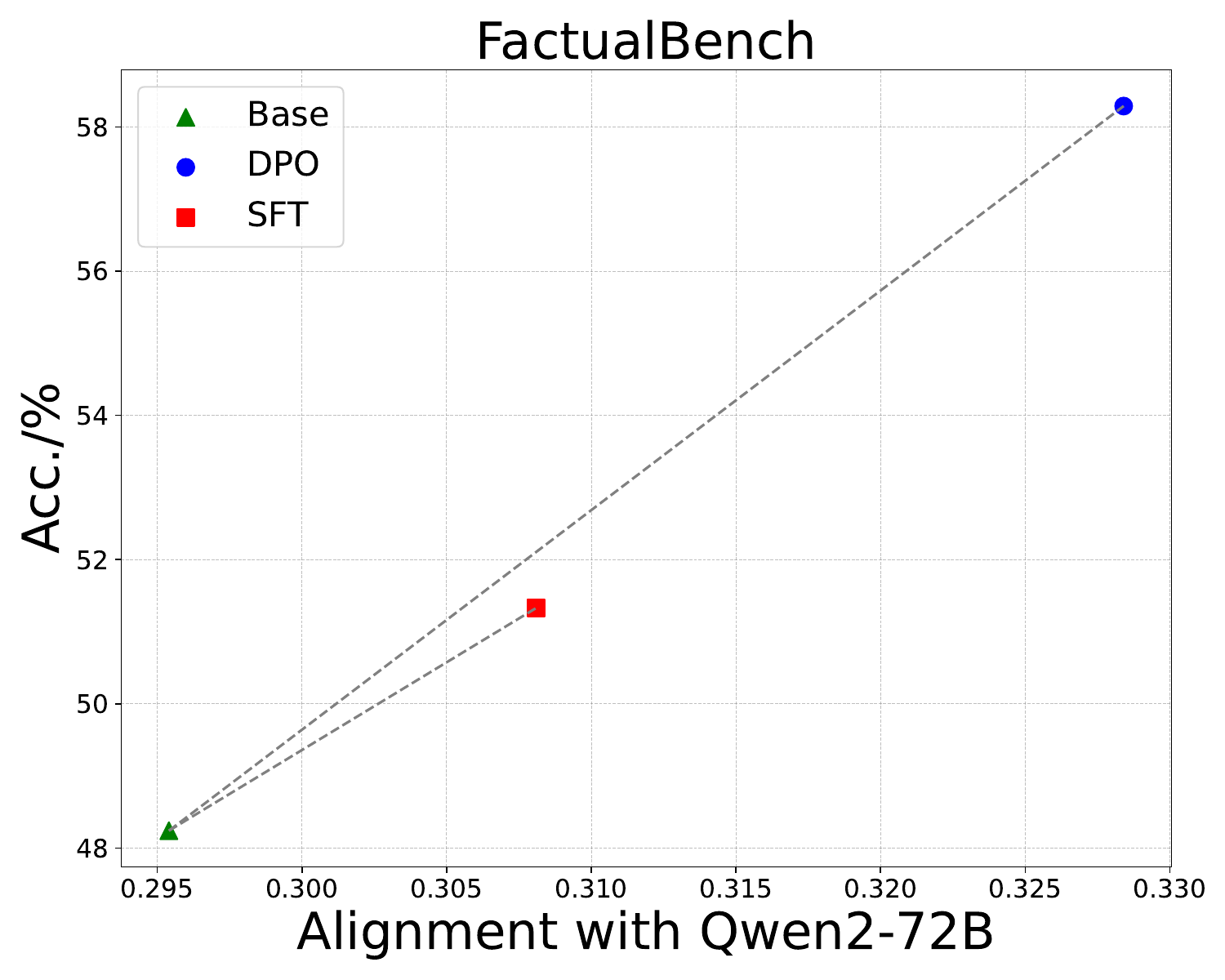} \hfill
  \includegraphics[width=0.24\textwidth]{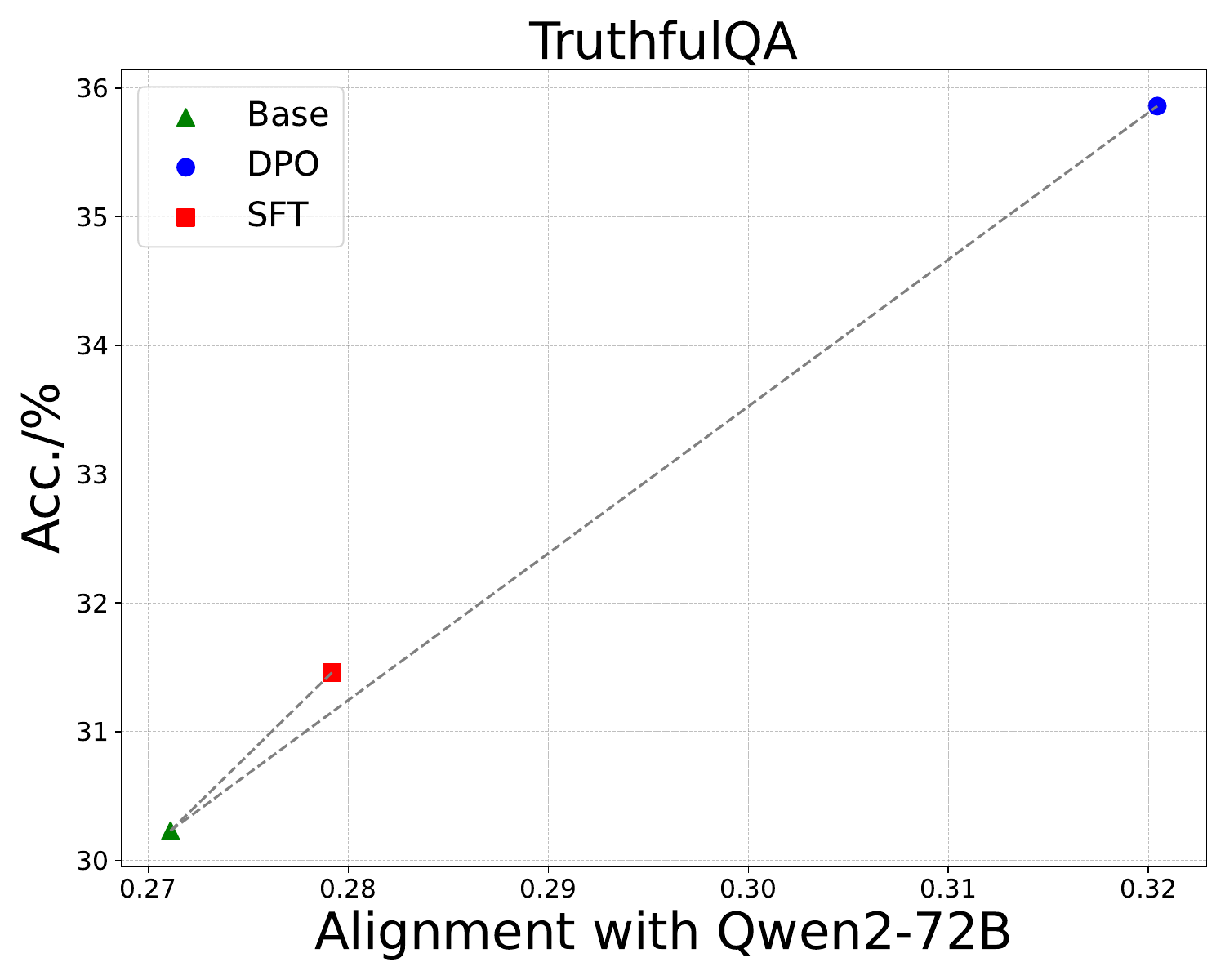} \hfill
  \includegraphics[width=0.24\textwidth]{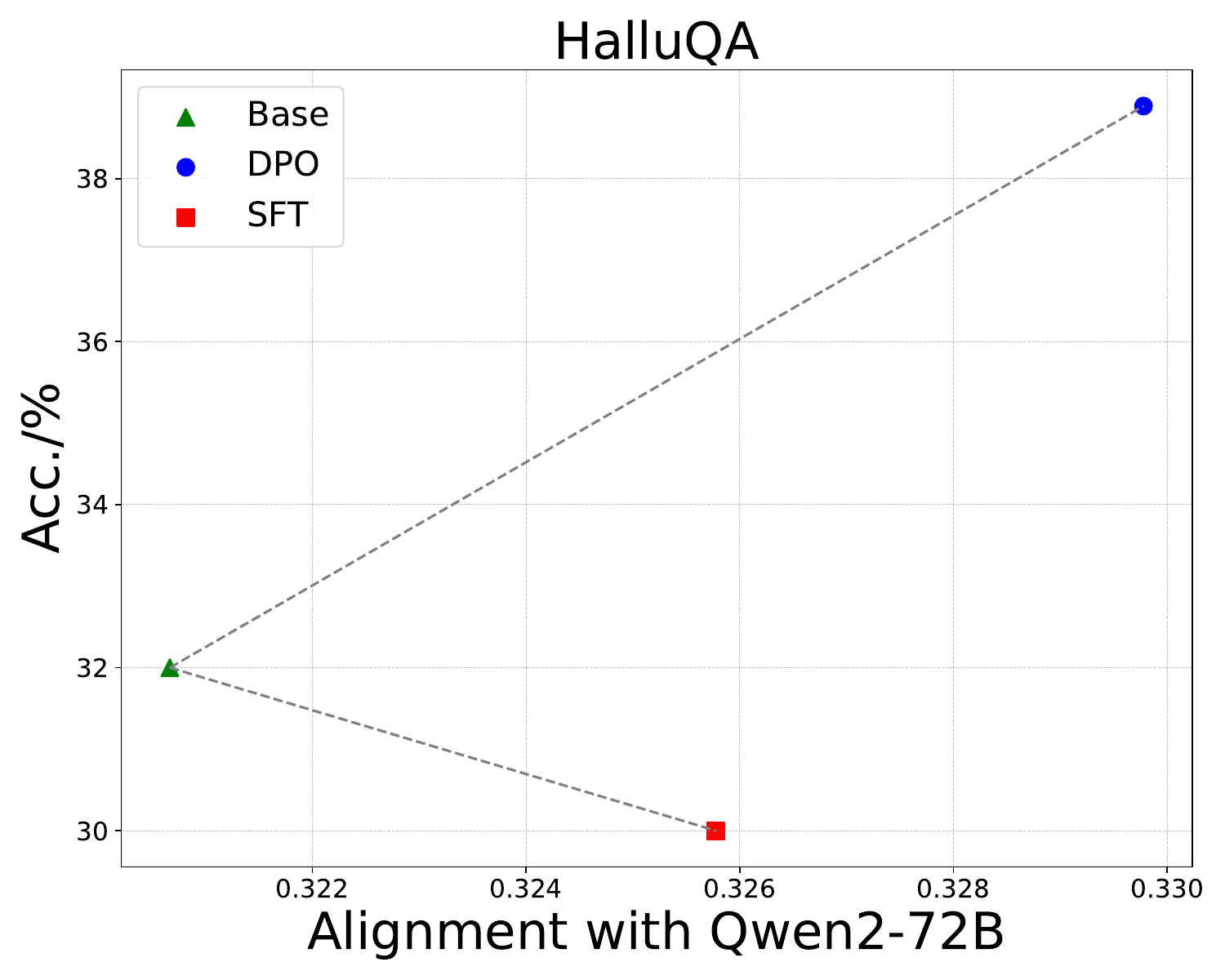} \hfill
  \includegraphics[width=0.24\textwidth]{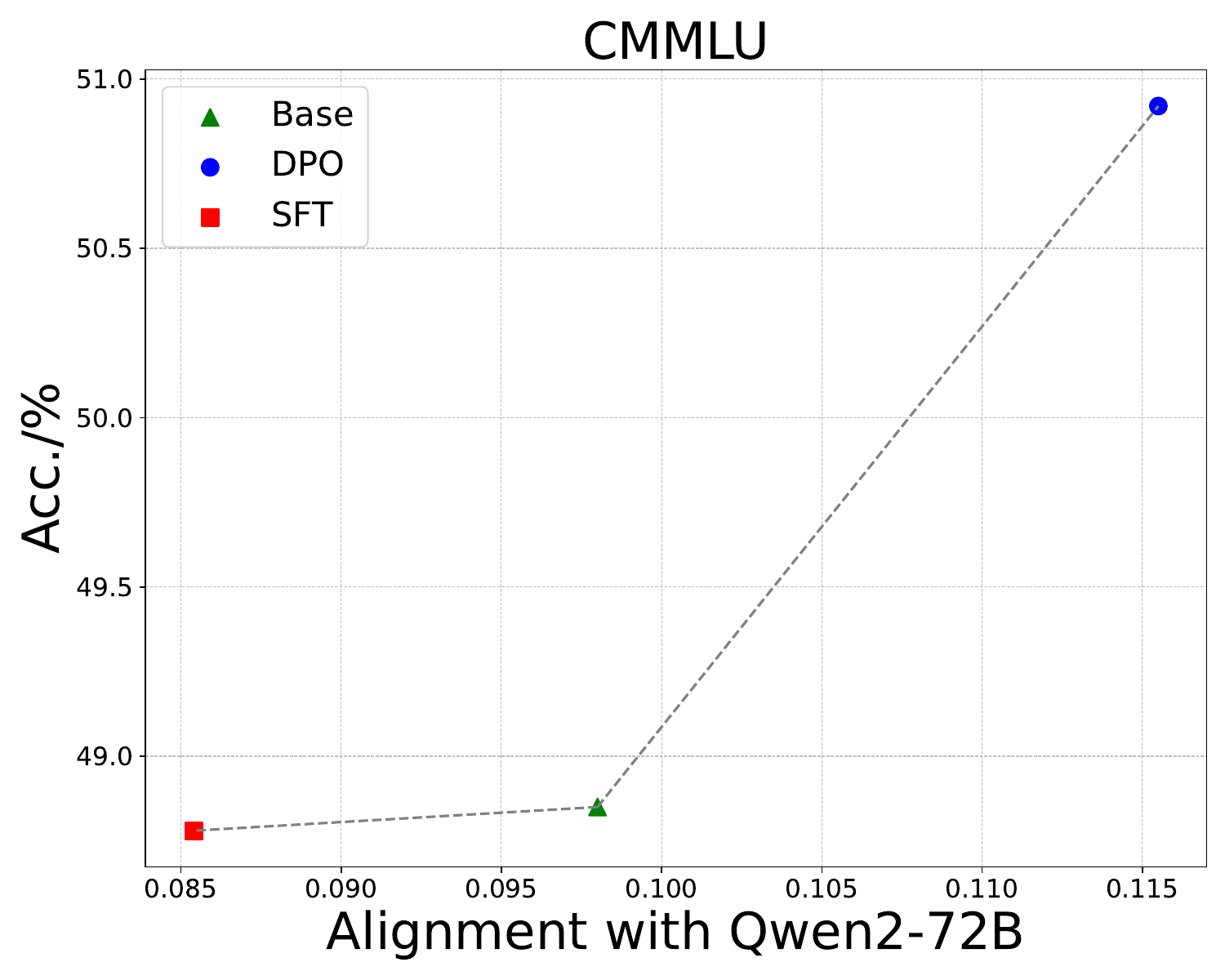}
  \caption {Changes of Baichuan1 alignment with Qwen2-72B-Instruct on four benchmarks after training.}
  \label{fig:alignment}
\end{figure*}

\begin{table}[!t]
  \centering
   \resizebox{\linewidth}{!}{%
  \begin{tabular}{c|cccc}
    \toprule[1.5pt]
    Loss & FactualBench & AlignBench & AlpacaEval & $\Delta$Avg. \\
    \midrule
    \multicolumn{5}{c}{\sc Baichuan1-Chat} \\ 
    \midrule
    SFT (single label) & 51.33 & 5.04 & 37.58 & +1.29 \\
    SFT (all labels) & 52.37 & 5.03 & 31.06 & +0.32 \\
    \midrule
    DPO (small) & 57.37 & 5.30 & 54.84 & +3.90 \\
    DPO (full) & \bf 58.29 & \bf 5.38 & \bf 63.99 & \bf +4.97 \\
    \midrule
    SFT then DPO & 54.74 & 5.07 & 54.53 & +4.03 \\
    SFT + DPO & 57.16 & 5.13 & 63.91 & +4.09 \\
    \bottomrule[1.5pt]
  \end{tabular}
  }
  \caption{Results after training on different losses.}
  \label{tab:loss ablation}
\end{table}

Furthermore, we argue that models obtain better representation ability after DPO. \citet{huh2024platonic} have found that the representation alignment degree, measured by mutual nearest-neighbor metric, which we introduce its definition and calculation in Appendix \ref{cha:metric}, increases with performance. We calculate the Baichuan1 alignment with Qwen2-72B-Instruct \citep{qwen2}, which serves as a strong representation function, on several benchmarks and present the results in Figure \ref{fig:alignment}. The DPO model achieves higher accuracy and deeper alignment with Qwen2-72B than both the base and SFT models, indicating that a better representation ability is achieved.

\begin{figure}[!t]
\centering
  \includegraphics[width=\columnwidth]{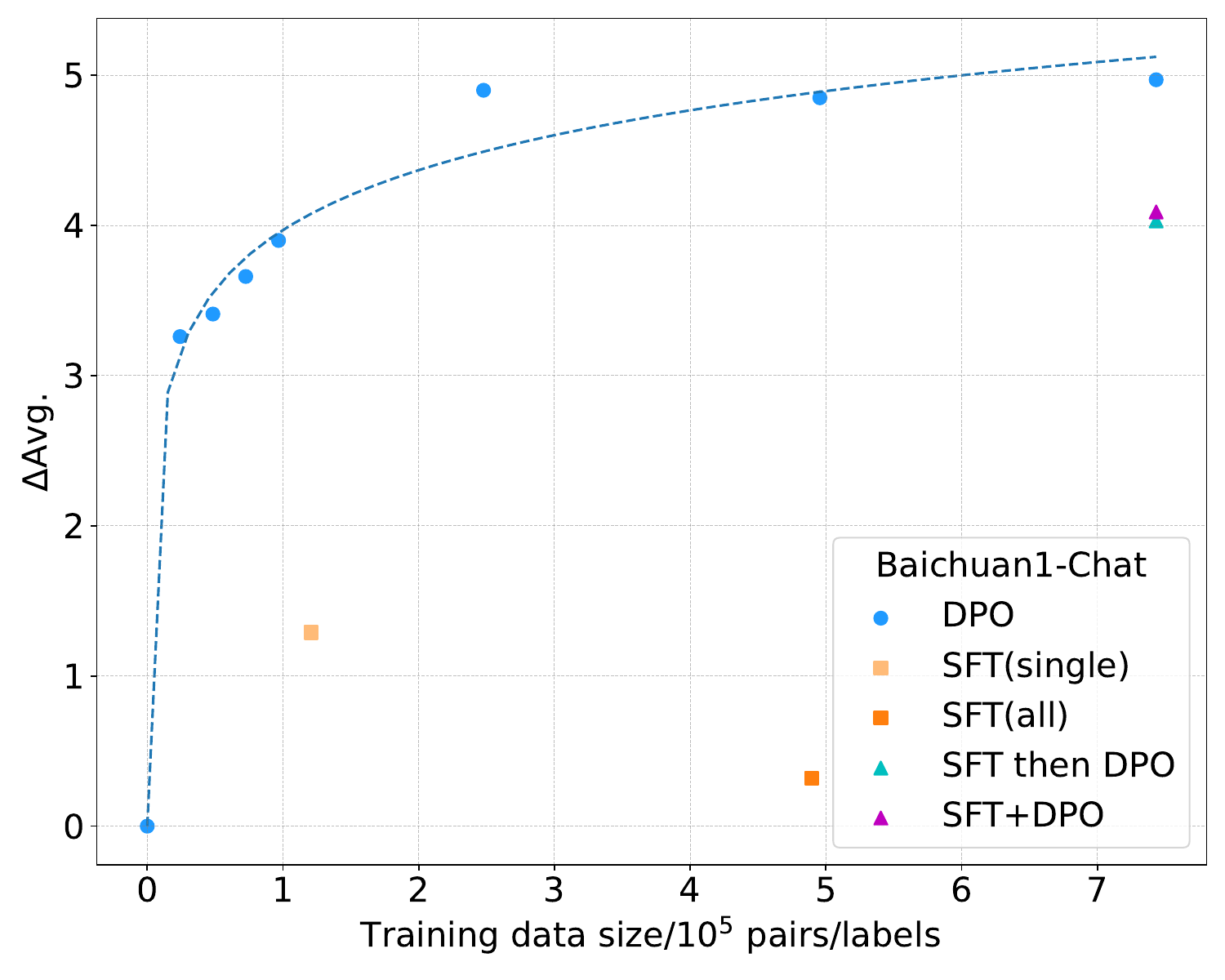}
  \caption{Baichuan1 performance improvement after DPO increases at a logarithmic rate with the training data size expanding.}
  \label{fig:scalinglaw}
\end{figure}

\textbf{Ablation on tuning data sizes.} A noticeable performance gap exists between the model trained on \textit{small} split and the one trained on \textit{full} split, motivating an exploration of the training efficacy of different tuning data sizes. We present the overall improvement of Baichuan1, measured by \textit{$\Delta$Avg.}, after DPO on different volumes of training questions in Figure \ref{fig:scalinglaw}. The improvement continues to increase (in logarithmic rate) as the size of preference pairs expands, stressing the benefit of a larger dataset, while early training with our method already improves the overall performance effectively.

%% file: sec/5_conclusion_etc.tex
\section{Conclusion}
We propose PKUE to mitigate factual hallucinations and achieve generalized improvement. Precise and simple factual QA is selected as our training task, and we align models on self-generated preference data to enhance the model's ability to utilize its knowledge. A large-scale, multi-domain Chinese dataset FactualBench is constructed from the Internet encyclopedia for training and evaluation. Extensive experiments demonstrate that PKUE significantly improves model performance across diverse tasks with the same and different languages, concerning factuality, helpfulness, and comprehensive advanced skills, which suggest that simply training on precise factual QA task has the potential for the overall improvement of the model.

\section{Acknowledgment}
This work was supported in part by NSFC Projects under Grant 92370124, Grant 92248303, Grant 62276149, Grant 62350080, Grant 62061136001, and Grant 62076147, in part by BNRist under Grant BNR2022RC01006, in part by Tsinghua Institute for Guo Qiang, and in part by the High Performance Computing Center, Tsinghua University.

\FloatBarrier

\section*{Limitations}
Although extensive experiments and ablation studies across diverse benchmarks validate the effectiveness of our method, certain limitations require further improvement.

\textbf{Alignment with more training algorithms.} The improvement curve observed in Figure \ref{fig:scalinglaw} exhibits an approximate logarithmic growth with diminishing marginal returns, and the model gains half of the improvement during the early training period. This suggests a potential in our training dataset for yielding further enhancement with thorough exploitation, such as adopting algorithms that are closer to online learning, including Proximal Policy Optimization \citep{schulman2017proximal} and iterative DPO algorithms \citep{iterativedpo, iterdpo2}.

\textbf{Hallucination mitigation in broader contexts.} Factual hallucinations occur not only in closed-book tasks, as discussed in this paper, but also in open-book tasks. These include text reading comprehension and text summarization tasks, which require the model's utilization of knowledge within the provided context instead of the model's existing knowledge. Investigations on more different tasks can verify whether the improvement on the model derived from our method can have a broader generalization.

\section*{Ethics Statement}
All experiments and analyses in this study are conducted for research purpose, aiming to enhance the factuality, robustness, and trustworthiness of LLMs and mitigate factual hallucinations. We collect data from the Internet following their license and only for research use.

The data source we use to build FactualBench is a publicly available Internet encyclopedia, which may contain information related to specific individuals, places, sensitive physiological or medical content. Yet, all the information is well-known, and we extract it without the intention to violate privacy or safety policies. Despite our efforts to ensure higher quality, the dataset could still contain inaccuracies or outdated information, which means that it should not be considered a golden knowledge base in any case and should only be adopted for research purposes.

The other benchmarks in this study are well-established, and we use them to assess the capabilities of different models and methods in line with their original purpose.

%% file: sec/X_appendix.tex
\newpage

\section{More Details of FactualBench}
In this section, we will introduce more details of our dataset, FactualBench, including prompts used for generation and evaluation, more examples of data in FactualBench, LLMs performance on FactualBench and related analyses.

\subsection{Prompts in Construction}
\label{cha:generation prompts}
Figure \ref{fig:generation prompt} shows the complete prompt that we use in \textit{Question Generation} stage, and Figure \ref{fig:generation prompt english} shows an English translation version. We include two manually written examples as few-shots and insert the target object with its description in the position marked in \textcolor{orange}{orange}.

Figure \ref{fig:filter prompt} shows the complete prompt we use in \textit{Question Filtering} stage, and Figure \ref{fig:filter prompt english} shows an English translation version. We list several types of low-quality cases and require GPT4 to judge whether the question falls into one. The question under judgment should be placed in the position marked in \textcolor{red}{red}.

\subsection{Prompt in Evaluation}
\label{cha:evaluation prompts}
Figure \ref{fig:eval prompt} shows the complete prompt that we use when evaluating the correctness of a response, and Figure \ref{fig:eval prompt english} shows an English translation version. We include five judging examples that cover the situations of answering correctly, answering incorrectly, and refusing to answer. The verifier is supposed to show its analysis before providing the judgment. The question, standard answer, and model's candidate answer should be placed in the position marked in \textcolor{green}{green}.

\subsection{More Examples in FactualBench}
\label{cha:more examples}
We list one example of each domain (exclude \textit{others}) in FactualBench in Table \ref{tab:more example1}, \ref{tab:more example2}, \ref{tab:more example3}. We provide English translations for reference only, and the questions are highlighted in \textcolor{blue!90}{blue}.

\subsection{Training of the Classifier}
\label{cha:classifier obtaining}
We obtain the classifier model by fine-tuning a Baichuan-13B-Instruct model, which has already possessed sufficient foundational capabilities. For the training data, we first collect a large volume of real user queries in advance, which are then labeled combining human judgments with GPT annotations, structured into a three-level taxonomy to ensure both classification accuracy and coverage. The model is fine-tuned exclusively on this task, enabling it to achieve robust performance in classifying queries across hundreds of categories. 

\subsection{Detailed Benchmark Results}
\label{cha:detailed result}
We benchmark 14 LLMs on our FactualBench test set: Baichuan1 (closed), Baichuan2 (closed), Qwen1.5-7B-Chat (open), Qwen2-7B-Instruct (open), Llama-3-8B-Instruct (open), Baichuan3 (closed), Yi-34B-Chat (open), Command-R-35B (open), Llama-3-70B-Instruct (open), Qwen2-72B-Instuct (open), Baichuan4 (closed), Command-R-plus-104B (open), DeepSeek-v2-0628 MoE-236B (open), GPT4-0125-preview (closed), among which DeepSeek and GPT4 are queried from API and others are run locally. We use the recommended generation configuration and code on huggingface\footnote{\url{https://huggingface.co/}} to generate responses, and we set max-new-tokens and max-length configuration large enough to ensure that models can complete all their responses to questions.

\begin{figure*}[ht]
  \includegraphics[width=\textwidth]{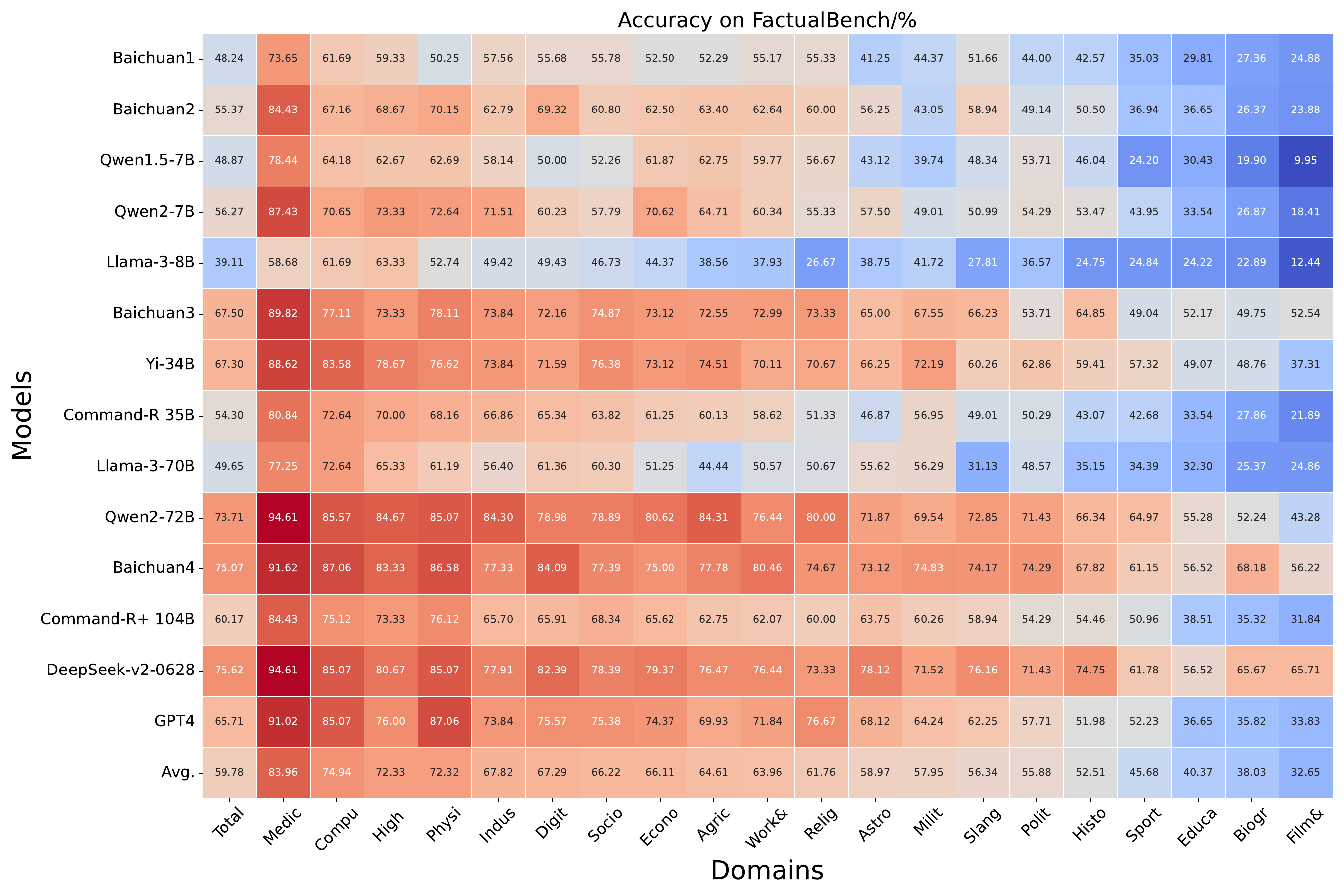} 
  \caption {Performance on FactualBench at domain level. We prioritize the chat / instruct version of these models.}
  \label{fig:heatmap benchmark}
\end{figure*}

We present the performance of 14 LLMs on our FactualBench at domain level using a heatmap in Figure \ref{fig:heatmap benchmark}. The first column presents the overall accuracy of the model, and the last line shows the average accuracy of all 14 models. We arrange domains from left to right in descending order of the average accuracy. Each domain is represented by its first five letters.

It is evident from the figure that as the number of model parameters increases, there is a corresponding upward trend in accuracy, while models with proficiency in Chinese demonstrate superior performance compared to those primarily proficient in English with approximate parameter numbers, which are aligned with expectations. Additionally, we have identified two key findings: 
\textbf{1) The performance of the same model can vary significantly across various domains}; 
\textbf{2) Different models share a consistency in relative ability on different domains}. Specifically, models tend to share similar domains where they achieve higher (or lower) accuracy, and there is no domain where one model excels (ranking in the top five accuracy domains) while another performs poorly (ranking in the bottom five accuracy domains). Interestingly, the film\&entertainment domain constitutes the largest portion of all data, but models exhibit the lowest accuracy on it among all domains.

We attribute the phenomenon to two possible primary factors. Firstly, the type of knowledge required varies across different domains. Secondly, the distribution of the pre-training data across these domains is uneven. These two factors contribute to the varying difficulty of tasks in different domains, and the differing levels of mastery that LLMs have over the knowledge pertinent to each domain, respectively.

\subsection{LLMs Responses in \textit{High Temp.}}
\label{cha:BO8 cases}
We present illustrative examples of model responses, including one instance from Baichuan1 on a test case (Table \ref{tab:bo1vsbo8 example bc}) and two examples from Qwen2-7B-Instruct on training cases (Table \ref{tab:bo8 example qwen}). For clarity, we provide English translations of key response details within square brackets ([]). The examples reveal that while Baichuan1 produces incorrect answers to questions under a low-temperature configuration, it can sometimes generate correct responses to them under a high-temperature configuration. Similarly, Qwen2 exhibits substantial variation in its responses under the high-temperature setting.

\begin{CJK*}{UTF8}{gbsn}
\begin{figure*}
\centering
\begin{tcolorbox}[title=Question generation, width=\textwidth]
我将提供给你一个对象和相关的参考文档，请针对对象提出最多\{提问个数：3\}个事实性问题。要求每个问题都具有唯一且准确的答案，避免答案模糊或存在争议，避免涉及主观判断的问题和时效性问题，要求答案可以在参考文档中直接找到。要求提问的问题表达清晰，问题中的名词指代明确，不需要依赖参考文档即可理解问题内容。对每个问题，给出 1 个标准答案和 3 个具有干扰性的错误答案。

下面是两个例子：

\vspace{1em}

【对象】：\{示例对象1\}

【参考文档】：\{关于示例对象1的百科内容简介\}

【问题1】：\{针对示例对象1提出的示例问题1\}

【标准答案】：\{示例问题1标准答案\}

【错误答案1】：\{示例问题1错误答案1\}

【错误答案2】：\{示例问题1错误答案2\}

【错误答案3】：\{示例问题1错误答案3\}

\vspace{1em}

【对象】：\{示例对象2\}

【参考文档】：\{关于示例对象2的百科内容简介\}

【问题1】：\{针对示例对象2提出的示例问题2\}

【标准答案】：\{示例问题2标准答案\}

【错误答案1】：\{示例问题2错误答案1\}

【错误答案2】：\{示例问题2错误答案2\}

【错误答案3】：\{示例问题2错误答案3\}

\vspace{1em}

对于以下的对象和参考文档，使用同样的格式生成问题、答案。

【对象】：\textcolor{orange}{\{对象：百科词条对象\}}

【参考文档】：\textcolor{orange}{\{文档：百科简介\}}

\vspace{1em}
\end{tcolorbox}
\caption {Prompt used to generate questions.}
\label{fig:generation prompt}
\end{figure*}
\end{CJK*}

\begin{figure*}
\centering
\begin{tcolorbox}[title=Question generation, width=\textwidth]
I will provide you with an object and its related reference description. Please generate up to \{Question number: 3\} factual questions about the object. Each question should have a unique and accurate answer, avoiding vague or contentious answers, subjective judgments, and time-sensitive. The answer should be directly found in the reference description. The question should be clearly expressed, with unambiguous noun references, and should not rely on the reference description for understanding. For each question, provide one standard answer and three misleading wrong answers.

Here are two examples:

\vspace{1em}

[Object]: \{Example Object 1\}

[Reference Description]: \{Brief introduction to Example Object 1\}

[Question 1]: \{Example question 1 related to Example Object 1\}

[Standard Answer]: \{Standard answer to Example question 1\}

[Wrong Answer 1]: \{Wrong answer 1 to Example question 1\}

[Wrong Answer 2]: \{Wrong answer 2 to Example question 1\}

[Wrong Answer 3]: \{Wrong answer 3 to Example question 1\}

\vspace{1em}

[Object]: \{Example Object 2\}

[Reference Description]: \{Brief introduction to Example Object 2\}

[Question 1]: \{Example question 2 related to Example Object 2\}

[Standard Answer]: \{Standard answer to Example question 2\}

[Wrong Answer 1]: \{Wrong answer 1 to Example question 2\}

[Wrong Answer 2]: \{Wrong answer 2 to Example question 2\}

[Wrong Answer 3]: \{Wrong answer 3 to Example question 2\}

\vspace{1em}

For the following object and reference description, generate questions and answers in the same format. 

[Object]: \textcolor{orange}{\{Object: Encyclopedia Entry Object\}}

[Reference Description]: \textcolor{orange}{\{Description: Encyclopedia Description\}}

\vspace{1em}
\end{tcolorbox}
\caption {Prompt used to generate questions (English translation).}
\label{fig:generation prompt english}
\end{figure*}

\begin{CJK*}{UTF8}{gbsn}
\begin{figure*}[ht]
\centering
\begin{tcolorbox}[title=Question filtering, width=\textwidth]
\textbf{User:}你是一个评估专家，下面需要你对一个问题的质量进行判断。

我会给你一个事实性知识问答问题, 你需要从以下几个方面分析这个问题，最终回答问题是【优质】还是【非优质】。

如果这个问题内存在代词指代不清，或无法明确理解问题含义，请回复【非优质】。

如果问题的答案不唯一，请回复【非优质】。

如果问题是时效性问题，且没有给出具体的背景时间点，请回复【非优质】。

如果问题没有以上情况，请回复【优质】。

请一步步思考，并在最后给出你的判断：【优质】或【非优质】。注意将你的最终判断写在中括号【】中！

\vspace{1em}

\textbf{Assistant:}明白了，我会按照你的要求和规则进行判断。

\vspace{1em}

\textbf{User:}问题是：

\textcolor{red}{\{待评价问题\}}

请给出你的判断：

\vspace{1em}
\end{tcolorbox}
\caption {Prompt used to filter out low-quality questions.}
\label{fig:filter prompt}
\end{figure*}
\end{CJK*}

\begin{figure*}[ht]
\centering
\begin{tcolorbox}[title=Question filtering, width=\textwidth]
\textbf{User: }You are an evaluation expert, and you need to assess the quality of a question.

I will provide you with a factual knowledge question, and you need to analyze the question from the following aspects to determine whether the question is of [High Quality] or [Low Quality].

If the question contains unclear pronoun references or cannot be clearly understood, please respond with [Low Quality].

If the answer to the question is not unique, please respond with [Low Quality].

If the question is time-sensitive and does not provide a specific time limitation, please respond with [Low Quality].

If none of the above situations apply, please respond with [High Quality].

Please think through the question step by step and give your final judgment as [High Quality] or [Low Quality]. Be sure to put your final judgment in square brackets []!

\vspace{1em}

\textbf{Assistant: }Understood. I will follow your requirements and rules for evaluation.

\vspace{1em}

\textbf{User: }The question is:

\textcolor{red}{\{Question to be evaluated\}}

Please provide your judgment:

\vspace{1em}
\end{tcolorbox}
\caption {Prompt used to filter out low-quality questions (English translation).}
\label{fig:filter prompt english}
\end{figure*}

\begin{CJK*}{UTF8}{gbsn}
\begin{figure*}[ht]
\centering
\begin{tcolorbox}[title=Answer evaluation, width=\textwidth]
给定一个问题以及对应的参考答案，根据参考答案和你掌握的知识，对候选答案是否回答正确进行评价。请注意，问题可能不具有唯一答案，此时只要候选答案言之成理即可。如果候选答案符合参考答案或言之成理，请回答【正确】；如果候选答案与参考答案矛盾或没有回答问题，请回答【错误】，并给出你的分析过程。下面是五个例子：

\vspace{1em}

【问题】：百川智能创始人王小川在什么时间与茹立云联合创立了该公司？

【参考答案】：百川智能创始人王小川于2023年4月10日与茹立云联合创立了该公司。

【候选答案】：王小川与茹立云于2023年4月共同创立了百川智能公司。

【评价】：根据参考答案，百川智能于2023年4月10日创立，候选答案认为是2023年4月创立，符合参考答案。【正确】

\vspace{1em}

【问题】：《采桑子·清明后三日作》是哪位诗人创作的？

【参考答案】：《采桑子·清明后三日作》是诗人龙榆生创作的。

【候选答案】：《采桑子·清明后三日作》是清代词人蒋春霖创作的一首词。

【评价】：根据参考答案，《采桑子·清明后三日作》是由诗人龙榆生创作，候选答案认为是蒋春霖创作，与参考答案矛盾。【错误】

\vspace{1em}

【问题】：李白的代表作有哪些？

【参考答案】：李白的代表作有《望庐山瀑布》《行路难》《蜀道难》《将进酒》《早发白帝城》《黄鹤楼送孟浩然之广陵》等。

【候选答案】：李白的代表作有《将进酒》《静夜思》《庐山谣》《早发白帝城》《赠汪伦》《望庐山瀑布》《行路难》《夜泊牛渚怀古》《登金陵凤凰台》《送友人》等。

【评价】：李白有许多代表作，答案不唯一，候选答案中的诗的确均为李白所写，言之成理。【正确】

\vspace{1em}

【问题】：哈蒂·温斯顿的主要作品有哪些？

【参考答案】：哈蒂·温斯顿的主要作品有《灵书妙探 第一季》。

【候选答案】：哈蒂·温斯顿（Hedy Lamarr）的主要作品有《Ecstasy》（1933年）， 《Algiers》（1938年）， 《Samson and Delilah》（1949年）等。

【评价】：哈蒂·温斯顿有许多作品，答案不唯一，但候选答案中的作品不是哈蒂·温斯顿的作品。【错误】

\vspace{1em}

【问题】：吴之番在哪次战斗中牺牲的？

【参考答案】：吴之番在清顺治二年八月二十六日的战斗中牺牲，这是嘉定三屠的一部分。

【候选答案】：对不起，我找不到关于“吴之番”的相关牺牲信息。这可能是因为您提供的信息有误或者该人物并不存在。

【评价】：根据参考答案，吴之番在顺治二年八月二十六日的战斗中牺牲，候选答案没有回答问题。【错误】

\vspace{1em}

下面是你需要评价的内容，请使用同样的格式给出评价。

【问题】：\textcolor{green}{\{问题\}}

【参考答案】：\textcolor{green}{\{参考答案\}}

【候选答案】：\textcolor{green}{\{候选答案\}}

【评价】：
\end{tcolorbox}
\caption {Prompt used to evaluate candidate answers to questions.}
\label{fig:eval prompt}
\end{figure*}
\end{CJK*}

\begin{figure*}[ht]
\centering
\small
\begin{tcolorbox}[title=Answer evaluation, width=\textwidth]
Given a question and its corresponding standard answer, evaluate whether the candidate answer correctly addresses the question based on the standard answer and your knowledge. Please note that the question may not have only one unique answer; in such cases, as long as the candidate answer is reasonable, it is acceptable. If the candidate answer aligns with the reference answer or is reasonable, please respond with [Correct]; if the candidate answer contradicts the reference answer or does not answer the question, please respond with [Incorrect], and provide your analysis. Here are five examples:

\vspace{1em}

[Question]: When did Wang Xiaochuan, the founder of Baichuan Inc., co-found the company with Ru Liyun?

[Standard Answer]: Wang Xiaochuan co-founded Baichuan Inc. with Ru Liyun on April 10, 2023.

[Candidate Answer]: Wang Xiaochuan and Ru Liyun co-founded Baichuan Inc. in April 2023.

[Evaluation]: According to the standard answer, Baichuan Inc. was founded on April 10, 2023. The candidate answer states it was founded in April 2023, which aligns with the reference answer. [Correct]

\vspace{1em}

[Question]: Which poet created "Cai Sang Zi · Qing Ming Hou San Ri Zuo"?

[Standard Answer]: "Cai Sang Zi · Qing Ming Hou San Ri Zuo" was created by the poet Long Yusheng.

[Candidate Answer]: "Cai Sang Zi · Qing Ming Hou San Ri Zuo" was created by the Qing Dynasty poet Jiang Chunlin.

[Evaluation]: According to the reference answer, "Cai Sang Zi · Qing Ming Hou San Ri Zuo" was created by Long Yusheng, while the candidate answer claims it was created by Jiang Chunlin, which contradicts the reference answer. [Incorrect]

\vspace{1em}

[Question]: What are the representative works of Li Bai?

[Standard Answer]: Li Bai's representative works include "Wang Lu Shan Pu Bu", "Xing Lu Nan", "Shu Dao Nan", "Qiang Jin Jiu", "Zao Fa Bai Di Cheng", and "Huang He Lou Song Meng Hao Ran Zhi Guang Ling", etc.

[Candidate Answer]: Li Bai's representative works include "Qiang Jin Jiu", "Jing Ye Si", "Lu Shan Yao", "Zao Fa Bai Di Cheng", "Zeng Wang Lun", "Wang Lu Shan Pu Bu", "Xing Lu Nan", "Ye Bo Niu Zhu Huai Gu", "Deng Jin Ling Feng Huang Tai", and "Song You Ren", etc.

[Evaluation]: Li Bai has many representative works, and the answer is not unique. The poems listed in the candidate answer are indeed all written by Li Bai, which is reasonable. [Correct]

\vspace{1em}

[Question]: What are the main works of Hattie Winston?

[Standard Answer]: Hattie Winston's main work is "Castle" (Season one).

[Candidate Answer]: Hedy Lamarr's main works include "Ecstasy" (1933), "Algiers" (1938), and "Samson and Delilah" (1949), etc.

[Evaluation]: Hattie Winston has many works, and the answer is not unique. However, the works listed in the candidate answer are not by Hattie Winston. [Incorrect]

\vspace{1em}

[Question]: In which battle did Wu Zhifan sacrifice?

[Standard Answer]: Wu Zhifan was sacrificed in the battle on August 26, the second year of the Shunzhi reign, which was part of the Jiadin Santu.

[Candidate Answer]: Sorry, I cannot find any information related to Wu Zhifan's sacrifice. This may be due to incorrect information you provided or because this person does not exist.

[Evaluation]: According to the standard answer, Wu Zhifan was sacrificed in the battle on August 26,  the second year of the Shunzhi reign, but the candidate answer did not answer the question. [Incorrect]

\vspace{1em}

Here is the content you need to evaluate, and please use the same format to provide your evaluation.

[Question]: \textcolor{green}{\{Question\}}

[Standard Answer]: \textcolor{green}{\{Standard Answer\}}

[Candidate Answer]: \textcolor{green}{\{Candidate Answer\}}

[Evaluation]: 

\end{tcolorbox}
\caption {Prompt used to evaluate candidate answers to questions (English translation).}
\label{fig:eval prompt english}
\end{figure*}

\begin{CJK*}{UTF8}{gbsn}
\begin{table*}[ht]
  \centering
    \resizebox{\textwidth}{!}{%
    \begin{tabular}{c|p{10cm}|p{10cm}}
    \toprule[1.5pt]
    \rowcolor{blue!10} Question & 韩国电影《人狼》是由哪位导演执导的？ & 河北师范大学最早起源于哪两所学校？ \\
    \rowcolor{blue!10} & Who directed the Korean movie 'Inrang'? & Which two schools did Hebei Normal University first originate from? \\ 
    \midrule
    Standard Answer & 电影《人狼》是由金知云执导的。 & 河北师范大学最早起源于顺天府学堂和北洋女师范学堂。 \\
    & The movie 'Inrang' is directed by Kim Jee-woon. & Hebei Normal University originated from Shuntianfu Official School and Beiyang Women's Normal School.  \\
    \midrule
    Wrong Answer1 & 电影《人狼》是由姜栋元执导的。 & 河北师范大学最早起源于河北师范学院和河北教育学院。\\
    & The movie 'Inrang' is directed by Kang Dong Won. & Hebei Normal University originated from Hebei Normal Institute and Hebei Institute of Education. \\
    \midrule
    Wrong Answer2 & 电影《人狼》是由韩孝周执导的。& 河北师范大学最早起源于河北职业技术师范学院和汇华学院。 \\
    & The movie 'Inrang' is directed by Han Hyo Joo. & Hebei Normal University originated from Hebei Vocational and Technical Normal College and Huihua College. \\
    \midrule
    Wrong Answer3 & 电影《人狼》是由郑雨盛执导的。 & 河北师范大学最早起源于北京大学和清华大学。\\
    & The movie 'Inrang' is directed by Jung Woo Sung. & Hebei Normal University originated from Peking University and Tsinghua University. \\ 
    \midrule
    Domain & 影视娱乐 & 教育培养 \\
    & film\&entertainment & education\&training \\
    \midrule[1.5pt]
    \rowcolor{blue!10} Question & 苯丙氨酸的化学式是什么？ & 谥号是在什么时期开始的？ \\
    \rowcolor{blue!10} & What is the chemical formula for phenylalanine? & When did posthumous titles begin? \\ 
    \midrule
    Standard Answer & 苯丙氨酸的化学式是C9H11NO2。 & 谥号始于西周。 \\
    & The chemical formula for phenylalanine is C9H11NO2. & The posthumous title began in the Western Zhou Dynasty.  \\
    \midrule
    Wrong Answer1 & 苯丙氨酸的化学式是C8H11NO2。 & 谥号始于东周。\\
    & The chemical formula for phenylalanine is C8H11NO2. & The posthumous title began in the Eastern Zhou Dynasty. \\
    \midrule
    Wrong Answer2 & 苯丙氨酸的化学式是C9H10NO2。& 谥号始于秦朝。 \\
    & The chemical formula for phenylalanine is C9H10NO2. & The posthumous title began in the Qin Dynasty. \\
    \midrule
    Wrong Answer3 & 苯丙氨酸的化学式是C9H11NO3。 & 谥号始于汉朝。\\
    & The chemical formula for phenylalanine is C9H11NO3. & The posthumous title began in the Han Dynasty. \\ 
    \midrule
    Domain & 数理化生 & 历史国学 \\
    &physics, chemistry, mathematics\&biology & history\&traditional culture \\
    \midrule[1.5pt]
    \rowcolor{blue!10} Question & 中国电影“第六代导演”之一王小帅的电影处女作是什么？ & 法律关系的构成要素有哪些？ \\
    \rowcolor{blue!10} & What is the debut film of Wang Xiaoshuai, one of the "sixth generation directors" of Chinese cinema? & What are the constituent elements of legal relationships?  \\ 
    \midrule
    Standard Answer & 王小帅的电影处女作是《冬春的日子》。 & 法律关系的构成要素有三项：法律关系主体，法律关系内容，法律关系客体。 \\
    & Wang Xiaoshuai's debut film is 'THE DAYS'. & There are three elements that make up a legal relationship: the subject of the legal relationship, the content of the legal relationship, and the object of the legal relationship.  \\
    \midrule
    Wrong Answer1 & 王小帅的电影处女作是《扁担姑娘》。 & 法律关系的构成要素有三项：法律关系主体，法律关系形式，法律关系客体。\\
    & Wang Xiaoshuai's debut film is 'So Close to Paradise'. & There are three elements that make up a legal relationship: the subject of the legal relationship, the form of the legal relationship, and the object of the legal relationship. \\
    \midrule
    Wrong Answer2 & 王小帅的电影处女作是《十七岁的单车》。& 法律关系的构成要素有三项：法律关系主体，法律关系内容，法律关系方式。 \\
    & Wang Xiaoshuai's debut film is 'Beijing Bicycle'. & There are three elements that make up a legal relationship: the subject of the legal relationship, the content of the legal relationship, and the method of the legal relationship. \\
    \midrule
    Wrong Answer3 & 王小帅的电影处女作是《青红》。 & 法律关系的构成要素有三项：法律关系主体，法律关系内容，法律关系目标。\\
    & Wang Xiaoshuai's debut film is 'Shanghai Dreams'. &  There are three elements that make up a legal relationship: the subject of the legal relationship, the content of the legal relationship, and the objective of the legal relationship. \\ 
    \midrule
    Domain & 人物百科 & 政治法律 \\
    & biography & politics\&law \\
    \bottomrule[1.5pt]
    \end{tabular}%
    }
  \caption{More examples in FactualBench (part 1).}
  \label{tab:more example1}
\end{table*}
\end{CJK*}

\begin{CJK*}{UTF8}{gbsn}
\begin{table*}[ht]
  \centering
    \resizebox{\textwidth}{!}{%
    \begin{tabular}{c|p{10cm}|p{10cm}}
    \toprule[1.5pt]
    \rowcolor{blue!10} Question & 国家金融监督管理总局是在哪一年揭牌的？ & MemCache是由谁开发的？ \\
    \rowcolor{blue!10} & In which year was the Chinese National Financial Supervisory Administration unveiled? & Who developed MemCache?  \\ 
    \midrule
    Standard Answer & 国家金融监督管理总局是在2023年揭牌的。 &  MemCache是由LiveJournal的Brad Fitzpatrick开发的。 \\
    & The Chinese National Financial Supervisory Administration was unveiled in 2023.& MemCache was developed by Brad Fitzpatrick from LiveJournal.  \\
    \midrule
    Wrong Answer1 & 国家金融监督管理总局是在2022年揭牌的。 & MemCache是由Facebook的Mark Zuckerberg开发的。\\
    & The Chinese National Financial Supervisory Administration was unveiled in 2022. & MemCache was developed by Mark Zuckerberg from Facebook. \\
    \midrule
    Wrong Answer2 & 国家金融监督管理总局是在2021年揭牌的。& MemCache是由Google的Larry Page开发的。 \\
    & The Chinese National Financial Supervisory Administration was unveiled in 2021. & MemCache was developed by Larry Page from Google. \\
    \midrule
    Wrong Answer3 & 国家金融监督管理总局是在2020年揭牌的。 & MemCache是由Microsoft的Bill Gates开发的。\\
    & The Chinese National Financial Supervisory Administration was unveiled in 2020. &  MemCache was developed by Bill Gates from Microsoft. \\ 
    \midrule
    Domain & 经济管理 & 计算机科学 \\
    & economics\&management & computer science \\
    \midrule[1.5pt]
    \rowcolor{blue!10} Question & 瑞舒伐他汀的主要作用部位是哪里？ & “枫丹白露”这个名字的原义是什么？ \\
    \rowcolor{blue!10} & What is the main site of action of rosuvastatin? & What is the original meaning of 'Fontainebleau'?  \\ 
    \midrule
    Standard Answer & 瑞舒伐他汀的主要作用部位是肝。 &  “枫丹白露”的法文原义为“美丽的泉水”。 \\
    & The main site of action of rosuvastatin is the liver. & The original French meaning of "Fontainebleau" is "beautiful spring water".  \\
    \midrule
    Wrong Answer1 & 瑞舒伐他汀的主要作用部位是心脏。 &“枫丹白露”的法文原义为“宏伟的宫殿”。\\
    & The main site of action of rosuvastatin is the heart. & The original French meaning of "Fontainebleau" is "magnificent palace". \\
    \midrule
    Wrong Answer2 & 瑞舒伐他汀的主要作用部位是肾脏。 & “枫丹白露”的法文原义为“狩猎的行宫”。 \\
    & The main site of action of rosuvastatin is the kidney. & The original French meaning of "Fontainebleau" is "hunting palace".  \\
    \midrule
    Wrong Answer3 & 瑞舒伐他汀的主要作用部位是胃。 & “枫丹白露”的法文原义为“古老的城堡”。\\
    & The main site of action of rosuvastatin is the stomache. &  The original French meaning of "Fontainebleau" is "ancient castle". \\ 
    \midrule
    Domain & 医学 & 社会人文 \\
    & medical & sociology\&humanity \\
    \midrule[1.5pt]
    \rowcolor{blue!10} Question & 竹笋原产于哪里？ & 更新世是由哪位地质学家创用的？ \\
    \rowcolor{blue!10} & Where do bamboo shoots originate from? & Which geologist named the Pleistocene epoch?  \\ 
    \midrule
    Standard Answer & 竹笋原产于中国。 &  更新世是由英国地质学家莱伊尔创用的。 \\
    & Bamboo shoots originate from China.  & The Pleistocene was named by British geologist Lyell.   \\
    \midrule
    Wrong Answer1 & 竹笋原产于日本。 &更新世是由英国地质学家福布斯创用的。\\
    & Bamboo shoots originate from Japan. & The Pleistocene was named by British geologist Forbes.  \\
    \midrule
    Wrong Answer2 & 竹笋原产于印度。 & 更新世是由美国地质学家莱伊尔创用的。 \\
    & Bamboo shoots originate from India. & The Pleistocene was named by American geologist Lyell. \\
    \midrule
    Wrong Answer3 & 竹笋原产于泰国。 & 更新世是由中国地质学家莱伊尔创用的。\\
    & Bamboo shoots originate from Thailand. & The Pleistocene was named by Chinese geologist Lyell. \\ 
    \midrule
    Domain & 农林牧渔 & 天文地理  \\
    & agriculture, forestry, fisheries\&allied industries & astronomy\&geography \\
    \bottomrule[1.5pt]
    \end{tabular}%
    }
  \caption{More examples in FactualBench (part 2).}
  \label{tab:more example2}
\end{table*}
\end{CJK*}

\begin{CJK*}{UTF8}{gbsn}
\begin{table*}[ht]
  \centering
    \resizebox{\textwidth}{!}{%
    \begin{tabular}{c|p{10cm}|p{10cm}}
    \toprule[1.5pt]
    \rowcolor{blue!10} Question & 新奥尔良鹈鹕队在哪一年正式宣布球队改名为鹈鹕队？ & 宾利汽车公司是在哪一年创办的？ \\
    \rowcolor{blue!10} & In which year did the New Orleans Pelicans officially announce their name change to the Pelicans? & In which year was BentleyMotors Limited founded?  \\ 
    \midrule
    Standard Answer & 新奥尔良鹈鹕队在2013年正式宣布球队改名为鹈鹕队。 &  宾利汽车公司是在1919年创办的。 \\
    & The New Orleans Pelicans officially announced their name change to the Pelicans in 2013.& BentleyMotors Limited was founded in 1919.  \\
    \midrule
    Wrong Answer1 & 新奥尔良鹈鹕队在2012年正式宣布球队改名为鹈鹕队。 & 宾利汽车公司是在1920年创办的。\\
    & The New Orleans Pelicans officially announced their name change to the Pelicans in 2012. & BentleyMotors Limited was founded in 1920. \\
    \midrule
    Wrong Answer2 & 新奥尔良鹈鹕队在2014年正式宣布球队改名为鹈鹕队。&宾利汽车公司是在1918年创办的。 \\
    & The New Orleans Pelicans officially announced their name change to the Pelicans in 2014. & BentleyMotors Limited was founded in 1918. \\
    \midrule
    Wrong Answer3 & 新奥尔良鹈鹕队在2015年正式宣布球队改名为鹈鹕队。 & 宾利汽车公司是在1921年创办的。\\
    & The New Orleans Pelicans officially announced their name change to the Pelicans in 2015. & BentleyMotors Limited was founded in 1921. \\ 
    \midrule
    Domain & 运动旅游 & 数码汽车 \\
    & sports\&tourism  & digital\&automotive \\
    \midrule[1.5pt]
    \rowcolor{blue!10} Question & 隔离开关主要用于什么？ & 鸦片战争是在哪一年开始的？ \\
    \rowcolor{blue!10} & What is the main use of disconnectors? & In which year did the Opium War begin?  \\ 
    \midrule
    Standard Answer & 隔离开关主要用于隔离电源、倒闸操作、用以连通和切断小电流电路。 &  鸦片战争是在1840年开始的。 \\
    & Disconnectors are mainly used for isolating power sources, switching operations, and connecting and disconnecting small current circuits. & The Opium War begin in 1840.  \\
    \midrule
    Wrong Answer1 & 隔离开关主要用于调节电压。 &鸦片战争是在1842年开始的。\\
    & Disconnectors are mainly used to regulate voltage. & The Opium War begin in 1842. \\
    \midrule
    Wrong Answer2 & 隔离开关主要用于转换电流。 & 鸦片战争是在1839年开始的。 \\
    & Disconnectors are mainly used to convert current. & The Opium War begin in 1839.  \\
    \midrule
    Wrong Answer3 & 隔离开关主要用于存储电能。 & 鸦片战争是在1841年开始的。\\
    & Disconnectors are mainly used for storing electrical energy. &  The Opium War begin in 1841. \\ 
    \midrule
    Domain & 工业工程  & 军武战争 \\
    & industrial engineering & military\&war \\
    \midrule[1.5pt]
    \rowcolor{blue!10} Question & 买了佛冷这个词是来源于哪首歌曲？ & 苏荷酒吧是在哪一年诞生的？ \\
    \rowcolor{blue!10} & What song does the meme 'Mai Le Fo Leng' come from? & In which year was Soho Bar founded?  \\ 
    \midrule
    Standard Answer & 买了佛冷这个词是来源于歌曲《I Love Poland》。 & 苏荷酒吧是在2003年诞生的。 \\
    & The meme 'Mai Le Fo Leng' comes from "I love Poland"  & Soho Bar was founded in 2003. \\
    \midrule
    Wrong Answer1 & 买了佛冷这个词是来源于歌曲《I Love China》。 &苏荷酒吧是在2000年诞生的。\\
    & The meme 'Mai Le Fo Leng' comes from "I love China" & Soho Bar was founded in 2000. \\
    \midrule
    Wrong Answer2 & 买了佛冷这个词是来源于歌曲《I Love America》。 & 苏荷酒吧是在2005年诞生的。 \\
    & The meme 'Mai Le Fo Leng' comes from "I love America" &Soho Bar was founded in 2005. \\
    \midrule
    Wrong Answer3 & 买了佛冷这个词是来源于歌曲《I Love England》。 &  苏荷酒吧是在2010年诞生的。\\
    & The meme 'Mai Le Fo Leng' comes from "I love England" &  Soho Bar was founded in 2010. \\ 
    \midrule
    Domain & 网词网梗 & 工作生活  \\
    & slang\&memes &  work\&life \\
    \midrule[1.5pt]
    \rowcolor{blue!10} Question & 视觉识别系统VI是什么的缩写？ & 风水业内公认的"龙脉之源"是哪里？ \\
    \rowcolor{blue!10} & What words is VI (a Vision System) abbreviation for? & Where is the recognized "source of dragon veins" in chinese feng shui?  \\ 
    \midrule
    Standard Answer & 视觉识别系统是Visual Identity的缩写。 &  风水业内公认的"龙脉之源"是昆仑山。 \\
    & VI abbreviation for Visual Identity. & The "source of dragon veins" in chinese feng shui is Kunlun Mountain.  \\
    \midrule
    Wrong Answer1 & 视觉识别系统是Visual Information的缩写。 & 风水业内公认的"龙脉之源"是长江。\\
    & VI abbreviation for Visual Information. & The "source of dragon veins" in chinese feng shui is Yangtze River. \\
    \midrule
    Wrong Answer2 & 视觉识别系统是Visual Interface的缩写。&风水业内公认的"龙脉之源"是黄河。 \\
    & VI abbreviation for Visual Interface. & The "source of dragon veins" in chinese feng shui is the Yellow River. \\
    \midrule
    Wrong Answer3 & 视觉识别系统是Visual Interaction的缩写。 & 风水业内公认的"龙脉之源"是太湖。\\
    & VI abbreviation for Visual Interface. &  The "source of dragon veins" in chinese feng shui is the Taihu Lake. \\ 
    \midrule
    Domain & 高新科技 & 信仰文化 \\
    & high technology  & religion\&culture \\
    \bottomrule[1.5pt]
    \end{tabular}%
    }
  \caption{More examples in FactualBench (part 3).}
  \label{tab:more example3}
\end{table*}
\end{CJK*}

\clearpage

\begin{CJK*}{UTF8}{gbsn}
\begin{table*}[ht]
  \centering
    \resizebox{\textwidth}{!}{%
    \begin{tabular}{c|p{20cm}|c}
    \toprule[1.5pt]
    \multicolumn{3}{c}{\sc Baichuan1-Chat} \\
    \midrule
    \rowcolor{blue!10} \multicolumn{3}{c}{Test split: 黄眉大王在《西游记》中最后是被谁降伏的？[标准答案：弥勒佛]} \\
    \rowcolor{blue!10} \multicolumn{3}{c}{Who ultimately subdued King Huangmei in \textit{Journey to the West?} [Standard Answer: \textbf{Maitreya Buddha}]} \\
    \midrule
    \small low temp. & 孙悟空 \textcolor{red}{[Sun WuKong]}。 & \ding{55} \\
    \midrule
    \small \multirow{3}{*}{high temp.} & 哪吒 \textcolor{red}{[Ne Zha]}$\backslash$n黄眉大王,第七十四回“长庚传报魔头狠 小圣战败母夜叉”中,红孩儿的叔叔被孙悟空、沙僧和护法伽蓝保着唐僧战败 & \ding{55} \\
    \cmidrule{2-3}
     & 孙悟空 \textcolor{red}{[Sun WuKong]}。第十九回$\backslash$n云栈洞悟空收八戒$\backslash$n浮屠山玄奘受心经 & \ding{55} \\
     \cmidrule{2-3}
     & 弥勒佛祖 \textcolor{green}{[Maitreya Buddha]}。黄眉大王是头目级妖怪,手下也有不少妖将,他趁孙悟空不在的时候捉了唐僧(其实是假冒的),后来弥勒佛祖亲自来了之后黄眉大王现出了原形。& \ding{51} \\
    \bottomrule[1.5pt]
    \end{tabular}%
    }
  \caption{Baichuan1 responses to a test case under low temperature and high temperature.}
  \label{tab:bo1vsbo8 example bc}
\end{table*}
\end{CJK*}

\begin{CJK*}{UTF8}{gbsn}
\begin{table*}[ht]
  \centering
    \resizebox{\textwidth}{!}{%
    \begin{tabular}{c|p{20cm}|c}
    \toprule[1.5pt]
    \multicolumn{3}{c}{\sc Qwen2-7B-Instruct} \\
    \midrule
    \rowcolor{blue!10} \multicolumn{3}{c}{Training split: 在《魔兽世界》中，牛头人在英语环境下被称为什么？[标准答案：Tauren]} \\
    \rowcolor{blue!10} \multicolumn{3}{c}{What is the shu'halo referred to as in English in \textit{World of Warcraft?} [Standard Answer: \textbf{Tauren}]} \\
    \midrule
    \small \multirow{3}{*}{high temp.} & 在《魔兽世界》这个虚构的系列游戏设定下，牛头人这一种族在其英语环境下被称为 \textcolor{green}{Tauren}。在该游戏背景故事中，牛头人们以他们的力量、荣耀概念的坚持和原生氏族形式的生活方式而著称，并被认为是艾泽拉斯这个主世界上的坚韧守护者之一。 & \ding{51} \\
    \cmidrule{2-3}
     & 在《魔兽世界》这个游戏中，牛头人种族在英语环境下被称为"\textcolor{red}{Ogres}"。但通常提到的牛头人应该指的是魔兽系列中的半神牛头人阵营，他们的正式名字叫作"High Hill Ogres"（山顶族）和"Grim Guor'gan Orges"（灰烬游侠），但也经常用"orcs"（食人魔）的统称来指代。这可能产生了一些混淆。正确而且具体的称呼应该是“Ogres”还是依据特定的情境下更加复杂的名称描述如 "Hill Gazelle" 或 "Mountain Goliaths" （具体根据《魔兽世界》的不同版本与时间而有所不同）。但在普通讨论或玩家交流中，“Ogres”是最常见的称呼方法。 & \ding{55} \\
     \cmidrule{2-3}
     & 在《魔兽世界》这个虚构的世界里，牛头人在这个英文环境下被称为\textcolor{red}{Trolls}。但在另一个文化背景中，特别是指真实生活中的非洲塞米诺尔人族裔时，他们被称为Touaregs或者Berbers，但这并不是《魔兽世界》中所使用的名称。在该游戏中，牛头人们属于暗夜 elves 的一种原住民，有自己的文化、语言和领袖，并在游戏的剧情中有很重要的地位。& \ding{55} \\
     \midrule
     \rowcolor{blue!10} \multicolumn{3}{c}{Training split: 哪个法甲球队在1992/93赛季获得了法甲球队历史上首个欧洲冠军联赛冠军？[标准答案：马赛]} \\
    \rowcolor{blue!10} \multicolumn{3}{c}{Which \textit{Ligue 1} football club won the first \textit{UEFA} Champions League championship} \\
     \rowcolor{blue!10} \multicolumn{3}{c}{in \textit{Ligue 1} history in the 1992/93 season? [Standard Answer: \textbf{Olympique de Marseille (OM)}]} \\
    \midrule
    \small \multirow{3}{*}{high temp.} & 在1992/93赛季获得法甲球队历史上首个欧洲冠军联赛（当时称为欧洲冠军杯）冠军的球队是马赛 \textcolor{green}{[OM]} 足球俱乐部。 & \ding{51} \\
    \cmidrule{2-3}
     & 在1992/93赛季，意大利的AC米兰队 \textcolor{red}{[A.C. Milan from Italy]} 在当年的欧洲冠军联赛中获得了历史性的首个欧洲冠军，而不是法甲球队。不过，有一支法甲球队在后来成功夺得了欧洲冠军联赛冠军，那便是法国球队马赛足球俱乐部，在1993/94赛季获得了其历史上首个也是至今唯一的欧洲冠军联赛冠军（以及后来所谓的"五冠王"）。 & \ding{55} \\
     \cmidrule{2-3}
     & 1992/93赛季获得欧洲冠军联赛（当时的赛事名称为欧洲冠军杯）冠军的是荷兰的阿贾克斯队 \textcolor{red}{[AFC Ajax from the Netherlands]} ，并非任何法甲球队。所以答案是：没有法甲球队在这个赛季获得过这个荣誉。& \ding{55} \\
    \bottomrule[1.5pt]
    \end{tabular}%
    }
  \caption{Qwen2-7B-Instruct responses to training cases under high temperature.}
  \label{tab:bo8 example qwen}
\end{table*}
\end{CJK*}

\FloatBarrier

\section{Training Details}
\label{sec:training details}
We complete all our training based on OpenRLHF \citep{hu2024openrlhf} framework, on 8 H800-80G NVIDIA GPUs, using AdamW optimizer \citep{loshchilov2017decoupled} and cosine scheduler. Learning rate is set to be $2e-6$ for SFT training, $1e-6$ for DPO training, and $1e-6$ DPO $+$ $1e-7$ SFT when fusing two loss functions together. Only $1$ epoch is trained on the tuning set, the batch size is set $128$, the DPO $\beta$ is set $0.1$ (which is consistent with the baselines \citep{flame, self-eval-skt}), and the weight decay is set $1e-4$ under all conditions. We list the sizes of all our tuning data in main experiment and ablation studies in Table \ref{tab:training data construction}. For SFT loss, a question is considered to be valid if the model correctly answers it at least one time; While for DPO loss, a question is considered to be valid if it receives both correct and incorrect answers from the model. As for baselines, we reproduce their methods following the settings in their papers.

\begin{table*}[ht]
  \centering
  \resizebox{0.8\textwidth}{!}{%
  \begin{tabular}{cccccc}
    \toprule[1.5pt]
    Loss & Split & Chosen & Rejected & \# Valid Questions & \# Labels/Pairs \\
    \midrule
    \multicolumn{6}{c}{\sc Qwen2-7B-Instruct} \\ 
    \midrule
    SFT & small & self & - & 16,845 & 16,845 \\
    SFT & small & Baichuan & - & 15,489 & 15,489 \\
    SFT & small & dataset & - & 24,000 & 24,000 \\
    DPO & small & self & self & 11,485 & 85,041 \\
    DPO\textsuperscript{1} & small & Baichuan & Baichuan & 12,949 & 96,737 \\
    DPO & small & dataset & dataset & 24,000 & 72,000 \\
    \midrule
    \multicolumn{6}{c}{\sc Baichuan1-Chat} \\ 
    \midrule
    SFT (single label) & full & self & - & 115,798 & 115,798 \\
    SFT (all labels) & full & self & - & 115,798 & 489,357 \\
    SFT & full & w/ desc. & - & 177,714 & 177,714 \\
    SFT & full & dataset & - & 177,714 & 177,714 \\
    DPO (small)\textsuperscript{1} & small & self & self & 12,949 & 96,737 \\
    DPO (full)\textsuperscript{2} & full & self & self & 98,805 & 743,333 \\
    DPO & full & w/ desc. & self & 177,714 & 881,932 \\
    DPO & full & dataset & self & 177,714 & 881,932 \\
    DPO & full & dataset & dataset & 177,714 & 533,142 \\
    SFT then DPO\textsuperscript{2} & full & self & self & 98,805 & 743,333 \\
    SFT + DPO\textsuperscript{2}  & full & self & self & 98,805 & 743,333 \\ 
    \bottomrule[1.5pt]
  \end{tabular}
  }
  \caption{Sizes of all our tuning data. Data with the same superscript \textsuperscript{1,2} are exactly the same.}
  \label{tab:training data construction}
\end{table*}

\section{Discussion on Verifier Accuracy}
\label{cha:verifier accuracy}
To quantify the accuracy of the verifier, we re-evaluate a partial set of $20k$ pairs from Qwen2-7B training data using GPT, which we have proved as a reliable evaluator in the main part of the paper. The evaluation reveals the following distribution: 9,276 pairs where GPT judges the chosen label to be better than the rejected label (chosen > rejected); 9,983 pairs where GPT judges the chosen label to be the same as the rejected label (chosen = rejected); and only 741 pairs where GPT judges the rejected label to be better than the chosen label (chosen < rejected). This indicates a very low rate of the \textit{wrong} pairs. In addition, we examine the judgment accuracy of single responses. Among 20,207 different answers within these $20k$ pairs, only 3,853 ($<20\%$) exhibit inconsistent evaluations between GPT and Qwen2.

To investigate the impact of this noise on DPO, we conduct three sets of training with each containing $10k$ pairs and trains for 5 epochs: High acc set (9,276 chosen > rejected + 724 chosen = rejected), Medium acc set (4,630 chosen > rejected + 5,000 chosen = rejected + 370 chosen < rejected), and Low acc set (2,259 chosen > rejected + 7,000 chosen = rejected + 741 chosen < rejected).

\begin{table*}[ht]
  \centering
  \resizebox{\textwidth}{!}{%
  \begin{tabular}{c|ccccc|c|c|c}
    \toprule[1.5pt]
     & FactualBench & TruthfulQA & HalluQA & CMMLU & HaluEval & AlignBench & AlpacaEval & $\Delta$ Avg. \\
    \midrule
    \multicolumn{9}{c}{\sc Qwen2-7B-Instruct} \\ 
    \midrule
   Base & 56.27 & 52.75 & 46.44 & 80.85 & 52.30 & 6.69 & 50.00 & - \\
   \midrule
   Low acc & 56.24 & 52.75 & 45.33 & 80.94 & 52.10 & 7.00 & 56.40 & +0.31 \\
   Medium acc & 56.79 & 52.39 & 45.56 & 81.10 & 52.39 & 7.06 & 56.93 & +0.55 \\
   High acc & 57.28 & 53.37 & 48.00 & 81.34 & 53.57 & 6.93 & 56.09 & +1.22 \\
    \bottomrule[1.5pt]
  \end{tabular}
  }
  \caption{Results after training on tuning sets with different accuracy.}
  \label{tab:acc experiment}
\end{table*}

As shown in Table \ref{tab:acc experiment}, higher pair accuracy generally improves training outcomes, leading to increased performance across nearly all benchmarks. Notably, even the model trained on the Low acc set outperforms the best baseline. This indicates that while enhanced evaluation accuracy holds greater potential for improvement, our method remains sufficiently effective even without a perfect verifier.

\section{Evaluation Details}
\label{cha:benchmark detail}
We choose six other open-source benchmarks in addition to FactualBench to evaluate the model's enhancement comprehensively. Models are required to respond to the questions or instructions in zero-shot condition and under default generation configuration. Official metrics are reported for all, and for model-based evaluation processes, we all choose GPT4 as evaluator.

\textbf{TruthfulQA} \cite{lin2022truthfulqa} is an English benchmark to measure whether a language model is truthful in generating answers. It contains 817 questions covering 38 domains. The questions are designed to cause imitative falsehoods which are due to a false belief or misconception. We use the generative part of TruthfulQA and adopt GPT4 to evaluate the response correctness, providing it with ground-truth correct answers and wrong answers.

\textbf{HalluQA} \citep{cheng2023evaluating} is a benchmark to measure hallucination in Chinese LLM. It contains 450 meticulously designed adversarial questions covering various domains to test imitative falsehoods of the model and factual knowledge. We use the generative part and its official prompt to evaluate the answer.

\textbf{CMMLU} \citep{li2023cmmlu} is a Chinese multiple-choice benchmark similar to MMLU \cite{MMLU}, comprising 67 topics with massive questions. We use the official script and code to evaluate the model's accuracy on the task.

\textbf{HaluEval} \citep{li2023halueval} is a large collection of generated and human-annotated English hallucinated samples to evaluate the performance of LLM on recognizing hallucinations. It is a discriminative task that requires the model to judge whether a response contains hallucination or not. We use the official prompt and only test on 10,000 samples from the QA part. The evaluation is based on string matching (e.g. ``Yes'' or ``No'') and if the model's judgment does not match any pattern, it will be considered as a wrong judgment.

\textbf{AlignBench} \citep{liu2023alignbench} is a Chinese benchmark for evaluating LLMs’ alignment skills. It contains 683 instructions on eight different tasks, including professional knowledge, mathematics, fundamental language ability, logical reasoning, advanced Chinese understanding, writing ability, task-oriented role play, and open-ended question. We use its official prompt format to evaluate answers in a model-based way.

\textbf{AlpacaEval} \citep{alpaca_eval} is a benchmark based on the AlpacaFarm \citep{dubois2023alpacafarm} evaluation set, which tests the model's instruction following ability. It contains 805 samples on different instructions, and calculates the winning rate against a base model. It has been used to indicate model's helpfulness in previous work \citep{flame}. In our experiments, the model before training is selected as the base model.

\section{Detailed Experiment Results}
In this section, we will provide more detailed results of the main experiment and ablation studies. Domain-level accuracy on FactualBench is presented in heatmaps, and performance on other benchmarks and sub-tasks of AlignBench is listed in tables.

\subsection{Main Experiments}
\label{cha:detailed main experiment result}
\begin{figure*}[ht]
  \includegraphics[width=\textwidth]{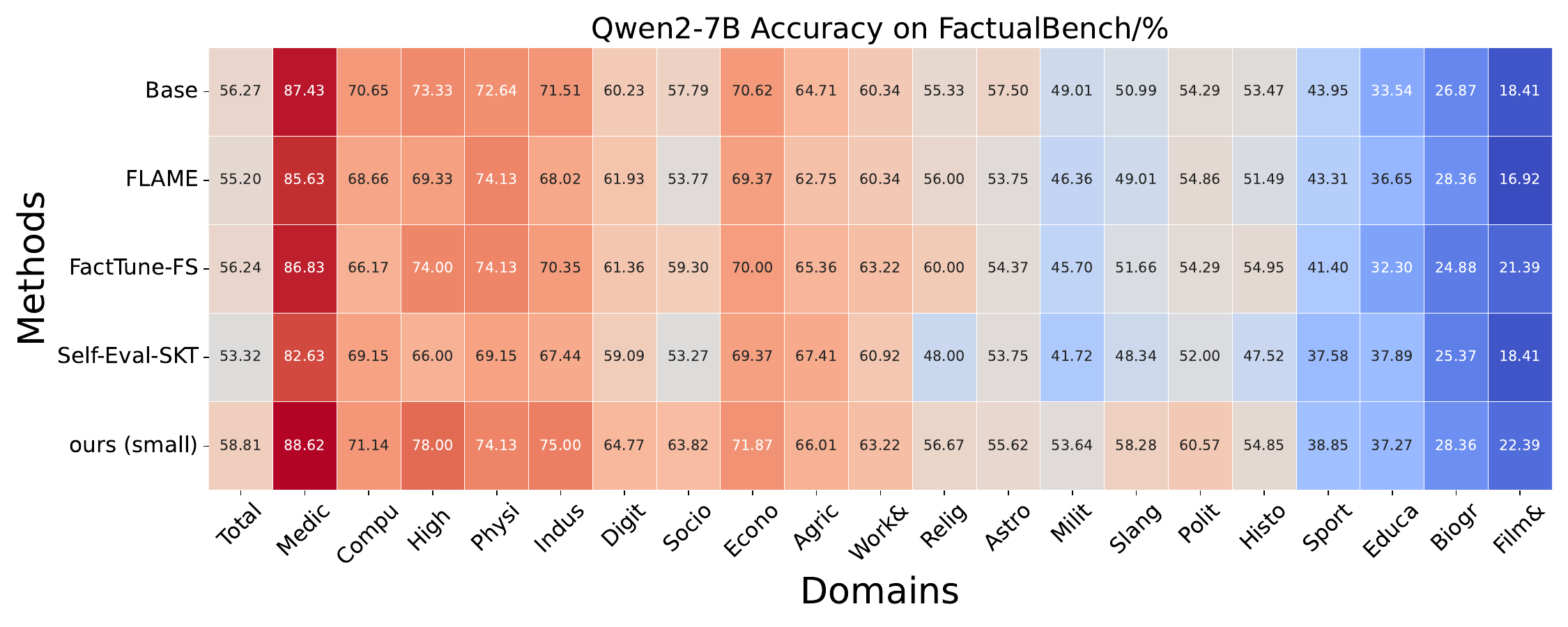} 
  \caption {Qwen2-7B-Instruct performance on FactualBench after different training methods.}
  \label{fig:heatmap qwen methods}
\end{figure*}

\begin{figure*}[ht]
  \includegraphics[width=\textwidth]{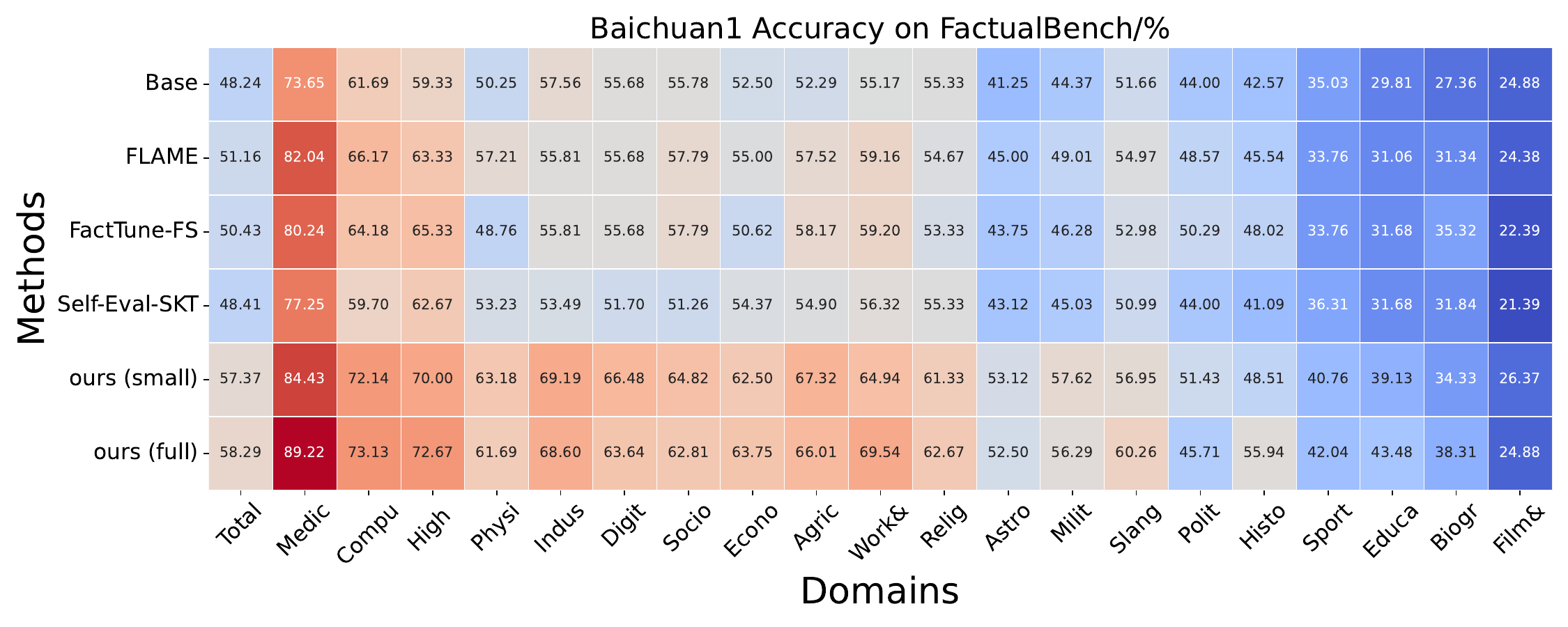} 
  \caption {Baichuan1-Chat performance on FactualBench after different training methods.}
  \label{fig:heatmap bc methods}
\end{figure*}

We present the performance of Qwen2-7B-Instruct and Baichuan1-Chat after training through our method and the other three baselines on FactualBench at domain-level in Figure \ref{fig:heatmap qwen methods} and Figure \ref{fig:heatmap bc methods}, respectively. The first column presents the overall accuracy of the model and we arrange the domains from left to right in the same order as in Figure \ref{fig:heatmap benchmark}. Each domain is represented by its first 5 letters.

\subsection{Ablation Studies}
\label{cha:detailed ablation experiment result}
\begin{figure*}[ht]
  \includegraphics[width=\textwidth]{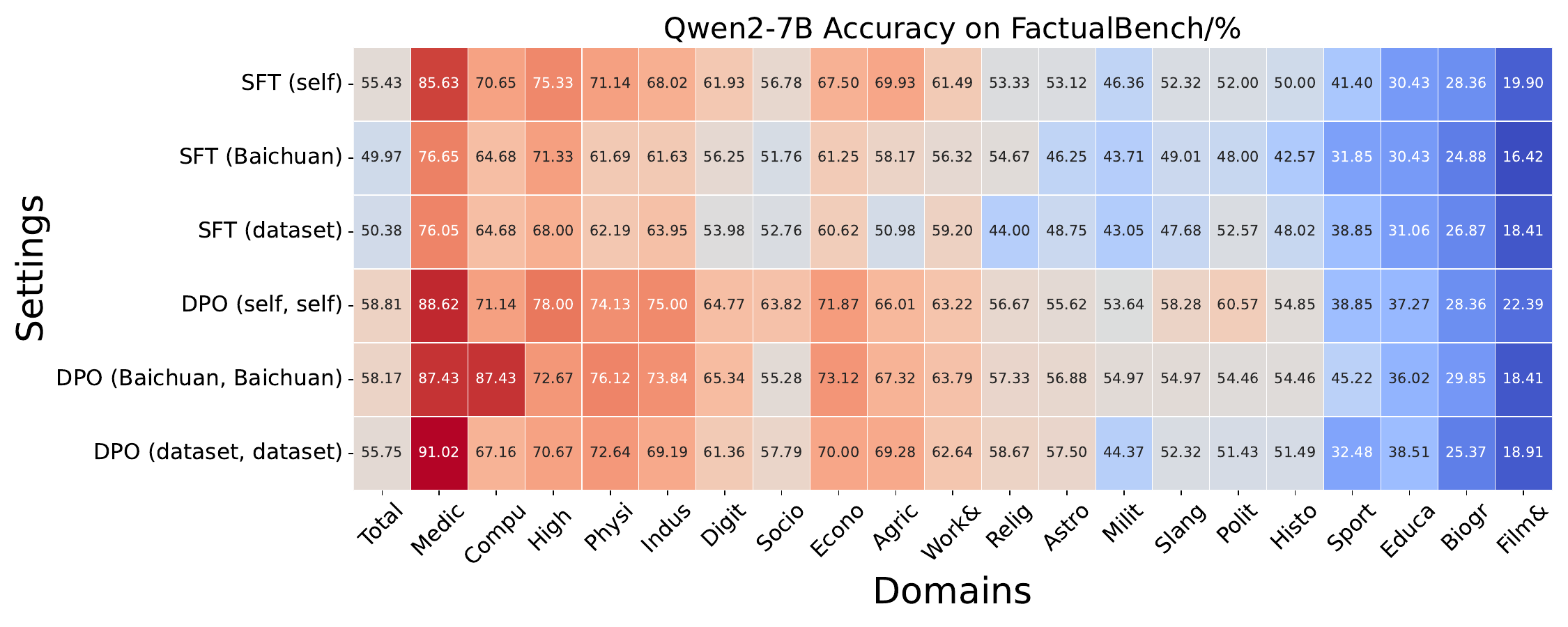} 
  \caption {Qwen2-7B-Instruct performance on FactualBench after training on different data sources.}
  \label{fig:heatmap qwen source}
\end{figure*}

\begin{figure*}[ht]
  \includegraphics[width=\textwidth]{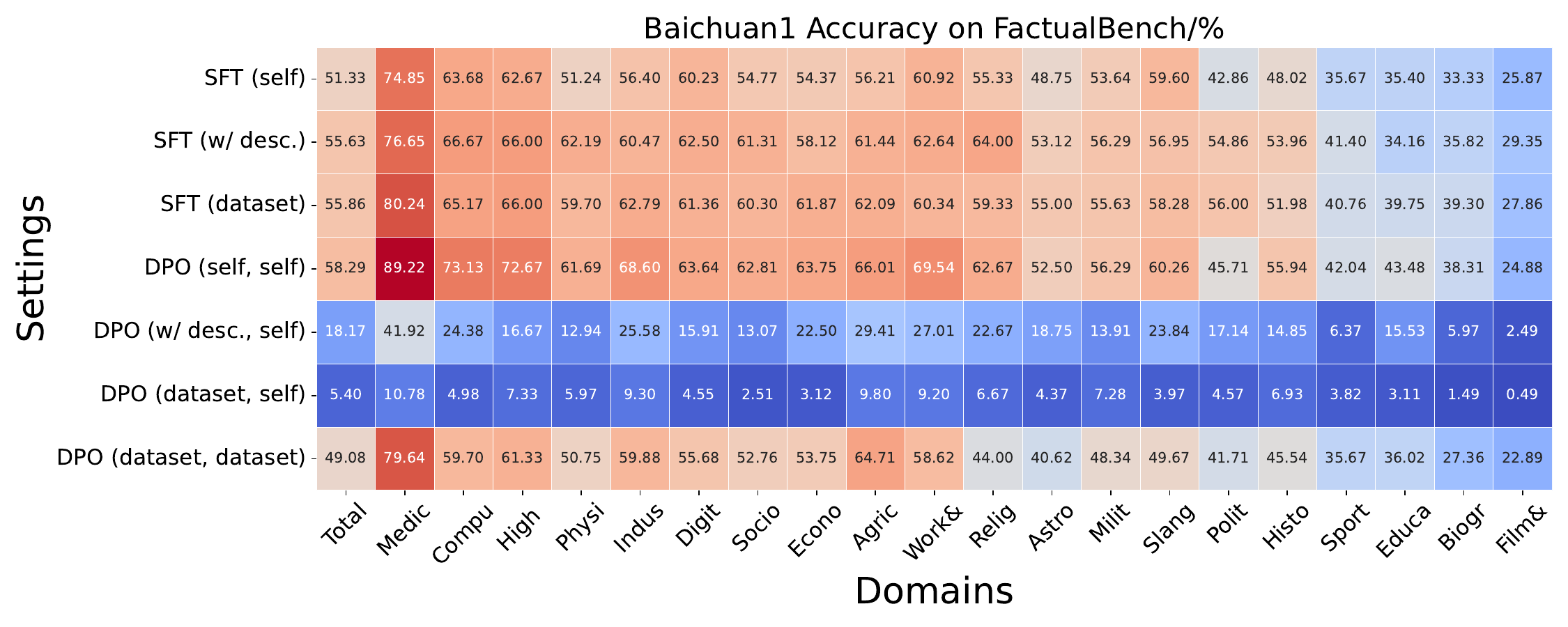} 
  \caption {Baichuan1-Chat performance on FactualBench after training on different data sources.}
  \label{fig:heatmap bc source}
\end{figure*}

For the ablation study on data sources, we present models performance on seven benchmarks in Table \ref{tab:data source ablation result}, on eight sub-tasks of AlignBench in Table \ref{tab:data source ablation result Alignbench}, and domain-level accuracy on FactualBench in Figure \ref{fig:heatmap qwen source}, Figure \ref{fig:heatmap bc source}. 

\begin{figure*}[ht]
  \includegraphics[width=\textwidth]{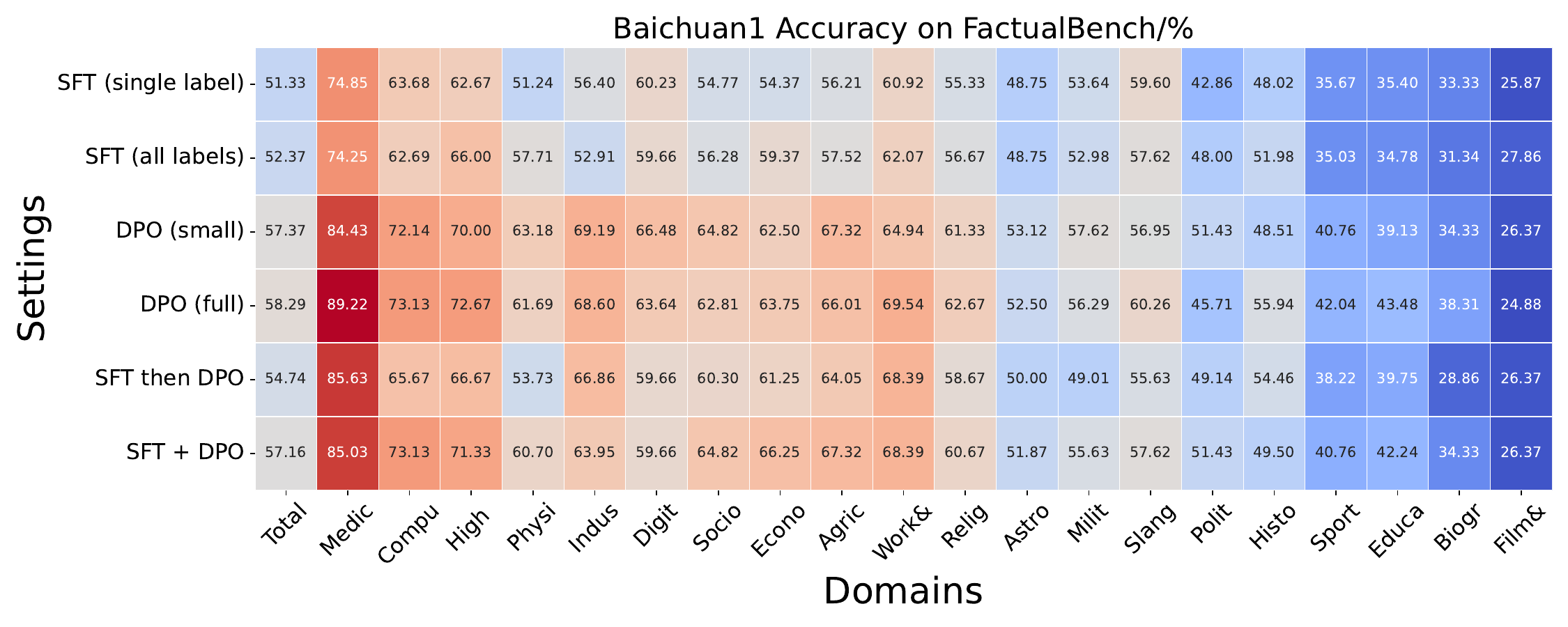} 
  \caption {Baichuan1-Chat performance on FactualBench after training using different loss functions.}
  \label{fig:heatmap bc loss}
\end{figure*}

For the ablation study on loss functions, we present Baichuan performance on seven benchmarks in Table \ref{tab:loss ablation result}, on eight sub-tasks of AlignBench in Table \ref{tab:loss ablation result Alignbench}, and domain-level accuracy on FactualBench in Figure \ref{fig:heatmap bc loss}. 

\begin{table*}[ht]
  \centering
   \resizebox{\textwidth}{!}{%
    \begin{tabular}{c|cc|ccccc|c|c|c}
    \toprule[1.5pt]
     Loss & Chosen & Rejected & FactualBench & TruthfulQA & HalluQA & CMMLU & HaluEval & AlignBench & AlpacaEval & $\Delta$Avg.\\ 
     \midrule
     \multicolumn{11}{c}{\sc Qwen2-7B-Instruct} \\ 
    \midrule
     SFT & self & - & 55.43 & 50.31 & 45.56 & 80.22 & 53.70 & 6.63 & 44.22 & -0.66  \\ 
     SFT & Baichuan & - & 49.97 & 29.87 & 24.67 & 77.49 & 42.05 & 4.98 & 15.03 & -13.61 \\
     SFT & dataset & - & 50.38 & 19.58 & 21.11 & 79.85 & 9.69 & 3.56 & 7.20 & -23.22  \\
     \midrule
     DPO & self & self & \bf 58.81 & \bf 54.47 & \bf 49.78 & \bf 82.15 & \bf 54.00 & \bf 6.96 & \bf 58.26 & \bf +2.22 \\ 
     DPO & Baichuan & Baichuan & 58.17 & 53.86 & 46.67 & 80.14 & 52.26 & 6.71 & 39.19 & +0.45  \\ 
     DPO & dataset & dataset & 55.75 & 52.14 & 46.22 & 80.77 & 51.70 & 6.50 & 36.06 & -0.65 \\ 
     \midrule
    \multicolumn{11}{c}{\sc Baichuan1-Chat} \\ 
    \midrule
     SFT & self & - & 51.33 & 31.46 & 30.00 & 48.78 & \bf 55.73 & 5.04 & 37.58 & +1.29  \\ 
     SFT & w/ desc. & -  & 55.63 & \bf 36.60 & 27.11 & 51.39 & 10.40 & 4.47 & 36.96 & -5.69  \\
     SFT & dataset & -  & 55.86 & 21.30 & 22.44 & 49.58 & 12.40 & 3.73 & 26.65 & -10.18  \\
     \midrule
     DPO & self & self & \bf 58.29 & 35.86 & \bf 38.89 & \bf 50.92 & 52.05 & \bf 5.38 & \bf 63.99 & \bf +4.97 \\ 
     DPO & w/ desc. & self & 18.17 & 13.10 & 9.33 & 48.05 & 48.57 & 4.07 & 32.80 & -13.67  \\ 
     DPO & dataset & self & 5.40 & 3.92 & 1.56 & 46.85 & 40.10 & 3.28 & 19.07 & -21.56  \\ 
     DPO & dataset & dataset & 49.08 & 28.89 & 19.78 & 50.70 & 54.89 & 4.82 & 39.07 & -1.40 \\ 
     \bottomrule[1.5pt]
    \end{tabular}
    }
  \caption{\label{tab:data source ablation result}
    Performance on seven benchmarks in data sources ablation study. We mark the best results in \textbf{bold}.
  }
\end{table*}

\begin{table*}[ht]
  \centering
   \resizebox{\textwidth}{!}{%
    \begin{tabular}{c|cc|cccccccc}
    \toprule[1.5pt]
     \multirow{2}{*}{Loss} & \multirow{2}{*}{Chosen} & \multirow{2}{*}{Rejected} & Professional & Mathe- & Fundamental & Logical & Advanced Chinese & Writing & Task-oriented & Open-ended \\ 
      & & & Knowledge & matics & Language Ability & Reasoning & Understanding & Ability & Role Play & Question \\
     \midrule
     \multicolumn{11}{c}{\sc Qwen2-7B-Instruct} \\ 
    \midrule
     SFT & self & - & \bf 6.74 & 6.40 & \bf 7.04 & 4.90 & 6.50 & 7.09 & 7.35 & 7.50  \\ 
     SFT & Baichuan & - & 5.26 & 3.72 & 5.88 & 3.41 & 5.31 & 5.60 & 5.59 & 6.42 \\
     SFT & dataset & - & 4.47 & 3.29 & 4.50 & 3.37 & 5.29 & 1.92 & 2.73 & 3.32 \\
     \midrule
     DPO & self & self & 6.63 & \bf 6.94 & 6.94 & \bf 5.56 & \bf 6.93 & \bf 7.43 & \bf 7.84 & \bf 7.92 \\ 
     DPO & Baichuan & Baichuan & 6.44 & 6.37 & 6.85 & 5.29 & 7.26 & 7.21 & 7.45 & 7.74 \\ 
     DPO & dataset & dataset & 6.10 & 6.36 & 6.76 & 4.70 & 6.59 & 7.23 & 7.64 & 7.07 \\ 
     \midrule
    \multicolumn{11}{c}{\sc Baichuan1-Chat} \\ 
    \midrule
     SFT & self & - & 5.78 & 2.59 & 5.47 & 3.30 & 5.66 & 6.11 & 6.25 & 6.58  \\ 
     SFT & w/ desc. & -  & 5.02 & 2.68 & 4.96 & 2.92 & 5.67 & 5.32 & 5.00 & 5.74  \\
     SFT & dataset & -  & 4.48 & 2.62 & 4.79 & 2.75 & 5.08 & 3.24 & 3.77 & 3.76 \\
     \midrule
     DPO & self & self & \bf 6.25 & \bf 3.03 & \bf 5.76 & \bf 3.55 & \bf 6.12 & \bf 6.52 & \bf 6.36 & \bf 6.79 \\ 
     DPO & w/ desc. & self & 3.62 & 1.93 & 4.88 & 2.63 & 4.47 & 5.81 & 5.53 & 5.34 \\ 
     DPO & dataset & self & 1.77 & 1.95 & 4.13 & 2.58 & 3.71 & 5.04 & 5.14 & 2.55  \\ 
     DPO & dataset & dataset & 4.67 & 2.60 & 5.53 & 3.30 & 5.50 & 6.40 & 6.17 & 6.00  \\ 
    \bottomrule[1.5pt]
    \end{tabular}
    }
  \caption{\label{tab:data source ablation result Alignbench}
    Performance on eight sub-tasks of AlignBench in data sources ablation study. We mark the best results in \textbf{bold}.
  }
\end{table*}

\begin{table*}[ht]
  \centering
   \resizebox{\textwidth}{!}{%
    \begin{tabular}{c|ccccc|c|c|c}
    \toprule[1.5pt]
     Loss & FactualBench & TruthfulQA & HalluQA & CMMLU & HaluEval & AlignBench & AlpacaEval & $\Delta$Avg.\\ 
     \midrule
    \multicolumn{9}{c}{\sc Baichuan1-Chat} \\ 
    \midrule
     SFT (single label)  & 51.33 & 31.46 & 30.00 & 48.78 & \bf 55.73 & 5.04 & 37.58 & +1.29  \\ 
     SFT (all labels)  & 52.37 & 28.76 & 26.44 & 50.15 & 53.90 & 5.03 & 31.06 & +0.32 \\ 
     \midrule
     DPO (small)  & 57.37 & 33.78 & 38.44 & 50.13 & 50.63 & 5.30 & 54.84 & +3.90 \\ 
     DPO (full)  & \bf 58.29 & 35.86 & \bf 38.89 & \bf 50.92 & 52.05 & \bf 5.38 & \bf 63.99 & \bf +4.97 \\ 
     \midrule
     SFT then DPO & 54.74 & \bf 37.33 & 36.67 & 50.72 & 54.02 & 5.07 & 54.53 & +4.03  \\ 
     SFT + DPO  & 57.16 & 34.76 & 38.22 & 50.78 & 52.31 & 5.13 & 63.91 & +4.09  \\ 
     \bottomrule[1.5pt]
    \end{tabular}
    }
  \caption{\label{tab:loss ablation result}
    Performance on seven benchmarks in loss functions ablation study. We mark the best results in \textbf{bold}.
  }
\end{table*}

\begin{table*}[ht]
  \centering
   \resizebox{\textwidth}{!}{%
    \begin{tabular}{c|cccccccc}
    \toprule[1.5pt]
     \multirow{2}{*}{Loss} & Professional & Mathe- & Fundamental & Logical & Advanced Chinese & Writing & Task-oriented & Open-ended \\ 
      & Knowledge & matics & Language Ability & Reasoning & Understanding & Ability & Role Play & Question \\
    \midrule
    \multicolumn{9}{c}{\sc Baichuan1-Chat} \\ 
    \midrule
     SFT (single label)  & 5.78 & 2.59 & 5.47 & 3.30 & 5.66 & 6.11 & 6.25 & 6.58  \\ 
     SFT (all labels)  & 5.46 & 2.88 & 5.60 & 3.25 & 5.57 & 6.19 & 6.17 & 6.63  \\ 
     \midrule
     DPO (small)  & 5.92 & 3.02 & 5.66 & 3.37 & 5.97 & \bf 6.53 & \bf 6.55 & 6.79  \\ 
     DPO (full)  & \bf 6.25 & \bf 3.03 & \bf 5.76 & \bf 3.55 & \bf 6.12 & 6.52 & 6.36 & 6.79  \\ 
     \midrule
     SFT then DPO & 5.57 & 2.66 & 5.53 & 3.01 & 6.00 & 6.33 & 6.32 & 6.92  \\ 
     SFT + DPO  & 5.60 & 2.79 & 5.57 & 3.16 & 6.05 & 6.17 & 6.41 & \bf 7.16 \\ 
     \bottomrule[1.5pt]
    \end{tabular}
    }
  \caption{\label{tab:loss ablation result Alignbench}
    Performance on eight sub-tasks of AlignBench in loss functions ablation study. We mark the best results in \textbf{bold}.
  }
\end{table*}

\section{More Experiments}
\label{cha: more experiments}
We provide more experiment results beyond the main experiment in this section.

\subsection{Experiments on More Models}
To demonstrate the effectiveness of our PKUE method on a wider range of models, we conduct experiments on Qwen2.5-7B-Instruct \citep{qwen2.5} and Llama3.1-8B-Instruct \citep{grattafiori2024llama} models. Both are the latest models released in 2024 and Llama3.1-8B is an English-proficient model.

We set the temperature=1.4 and top-p=0.9, top-k=50 that align with the main experiment settings, and train the models on the \textit{small} split. The verifier is set to be the same as the model to be trained. The results are shown in Table \ref{tab:experiment on more models}.

\begin{table*}[ht]
  \centering
   \resizebox{\textwidth}{!}{%
    \begin{tabular}{c|ccccc|c|c|c}
    \toprule[1.5pt]
     Model & FactualBench & TruthfulQA & HalluQA & CMMLU & HaluEval & AlignBench & AlpacaEval & $\Delta$Avg.\\ 
     \midrule
    \multicolumn{9}{c}{\sc Qwen2.5-7B-Instuct} \\ 
    \midrule
     Base  & 56.01 & 57.77 & 50.44 & 78.74 & 60.77 & 6.69 & 50.00 & -  \\ 
     \midrule
     PKUE (small)  & 58.52 & 59.61 & 54.67 & 80.00 & 63.28 & 6.81 & 54.91 & +2.26 \\
     \midrule
     \multicolumn{9}{c}{\sc Llama3.1-8B-Instuct} \\ 
     \midrule
     Base  & 33.94 & 50.55 & 12.89 & 55.67 & 65.48 & 3.98 & 50.00 & - \\ 
     \midrule
     PKUE (small)  & 43.01 & 52.30 & 17.78 & 56.72 & 66.55 & 4.81 & 52.83 & +4.35 \\ 
     \bottomrule[1.5pt]
    \end{tabular}
    }
  \caption{\label{tab:experiment on more models}
    PKUE performance on Qwen2.5-7B-Instruct and Llama3.1-8B-Instruct.
  }
\end{table*}

The observed results exhibit the same trends as in the main experiment. The application of PKUE yields consistent performance enhancement across all seven benchmarks for both models, achieving \textit{Avg.} improvement of 2.26 and 4.35, respectively. This further proves the effectiveness of PKUE. Moreover, the training effect on LLama3.1-8B underscores the utility of FactualBench for English-proficient models and suggests a deep relationship between the abilities of different languages.

\subsection{Comparisons with More Baselines}
To better confirm the superiority of PKUE and the poor generalizability of existing methods, we additionally examine two more decoding / inference strategies for factuality enhancement, which are widely compared in previous researches \citep{facttune, self-eval-skt}, Dola \citep{dola} and ITI \citep{li2023inference} on Llama3.1-8B-Instruct. We reproduce ITI and adopt the official implementation of Dola from the transformers library. The results are shown in Table \ref{tab:experiment on more baselines}.

\begin{table*}[ht]
  \centering
   \resizebox{\textwidth}{!}{%
    \begin{tabular}{c|ccccc|c|c|c}
    \toprule[1.5pt]
     Method & FactualBench & TruthfulQA & HalluQA & CMMLU & HaluEval & AlignBench & AlpacaEval & $\Delta$Avg.\\ 
     \midrule
     \multicolumn{9}{c}{\sc Llama3.1-8B-Instuct} \\ 
     \midrule
     Base  & 33.94 & 50.55 & 12.89 & 55.67 & 65.48 & 3.98 & 50.00 & - \\ 
     \midrule
     PKUE (small)  & \textbf{43.01} & \textbf{52.30} & 17.78 & \textbf{56.72} & \textbf{66.55} & \textbf{4.81} & \textbf{52.83} & +4.35 \\ 
     \midrule
     Dola  & \textcolor{red}{32.21} & \textcolor{red}{50.43} & 16.22 & 56.60 & \textcolor{red}{64.55} & 4.03 & \textcolor{red}{49.10} & +0.33 \\ 
     \midrule
     ITI  & \textcolor{red}{28.45} & 52.26 & \textbf{21.33} & \textcolor{red}{52.70} & \textcolor{red}{24.00} & \textcolor{red}{3.89} & \textcolor{red}{15.31} & \textcolor{red}{-6.78} \\ 
     \bottomrule[1.5pt]
    \end{tabular}
    }
  \caption{\label{tab:experiment on more baselines}
     Comparison between PKUE and more baselines on Llama3.1-8B-Instruct. We mark the decreased results in \textcolor{red}{red}, and the best results in \textbf{bold}.
  }
\end{table*}

Both methods indeed improve model performance on the targeted factual tasks like TruthfulQA and HalluQA. However, on most other factuality-concerned and beyond-factual benchmarks, these two baselines experience performance degradation and even severe drops, which showcases the advantage and significant effectiveness of our PKUE.

\subsection{Evaluations on More Benchmarks}
To further demonstrate the generalized improvement after PKUE training, we add three additional tasks, biography generation (BioGen) \citep{factscore}, SimpleQA \citep{wei2024measuring}, and HotpotQA \citep{yang2018hotpotqa} related to factuality, and evaluate the four selected models. 

\textbf{BioGen} is a task that requires LLM to generate people biographies, in the format of ``Tell me a bio of \texttt{entity}''. It is evaluated using the FActScore metric and reflects the factuality of LLMs in long-form open-ended tasks. We sample 100 celebrities from Wikipedia and report the average FActScore (\%).

\textbf{SimpleQA} is a benchmark that evaluates LLM factuality to short, fact-seeking questions. Compared with the other short-form factual tasks in our paper, SimpleQA is more challenging, as it is adversarially collected against GPT4 responses \citep{wei2024measuring}. We report the response accuracy (\%) in answering all 4,326 questions.

\textbf{HotpotQA} is a benchmark with multi-hop QA tasks. We use the distractor set to assess LLM performance under conditions with provided references (w/ ref) and without references (w/o ref). This approach allows us to test model's multi-hop reasoning capabilities both when contextual knowledge is available, and when it must rely solely on its internal knowledge. We report the response accuracy (\%) in answering all 7,405 questions.

\begin{table*}[ht]
  \centering
    \begin{tabular}{ccccc}
    \toprule[1.5pt]
     Model & BioGen & SimpleQA & HotpotQA w/ ref & HotpotQA w/o ref \\ \midrule
     Qwen2-7B & 50.8 & 3.44 & 75.72 & 39.38 \\  
     FLAME & 56.2 & 3.47 & - & -  \\ 
     FactTune-FS & \textbf{57.1} & \textcolor{red}{3.40} & - & - \\ 
     Self-Eval-SKT & 55.8 & \textcolor{red}{3.10} & - & - \\  
     PKUE (small) & 53.4 & \textbf{3.81} & 76.76 & 41.36 \\ \midrule
     Baichuan1 & 40.5 & 2.20 & 39.05 & 23.36 \\
     FLAME & \textbf{47.3} & 2.29 & - & - \\
     FactTune-FS & 45.5 & 2.29 & - & - \\
     Self-Eval-SKT & 46.1 & \textcolor{red}{2.03} & - & - \\
     PKUE (small) & 44.0 & \textbf{2.66} & 49.78 & 27.32 \\
     PKUE (full) & 46.3 & \underline{2.94} & - & - \\ \midrule
     Qwen2.5-7B & 56.4 & 3.68 & - & - \\
     PKUE (small) & \textbf{59.2} & \textbf{3.98} & - & - \\ \midrule
     Llama3.1-8B & 60.9 & 2.29 & - & - \\
     PKUE (small) & \textbf{63.9} & \textbf{5.92} & - & - \\
     Dola & 62.7 & 2.43 & - & -\\
     ITI & 61.3 & 2.50 & - & - \\
     \bottomrule[1.5pt]
   \end{tabular}%
  \caption{Experiments on more benchmarks: BioGen, SimpleQA, and HotpotQA. We mark the decreased results in \textcolor{red}{red}, and the best results in \textbf{bold}. Due to resource limitations, we do not conduct HotpotQA evaluation on all models.}
  \label{tab:experiment on more benchmarks}
\end{table*}

The experimental results in Table \ref{tab:experiment on more benchmarks} demonstrate that our PKUE can achieve competitive generalized improvement in long-form factual tasks, multi-hop factual tasks, and more difficult short-form precise QA tasks, even though baselines FLAME, FactTune-FS, and Self-Eval-SKT have in-domain training on the BioGen task. These results show that improvement on simple QA can generalize to other tasks with different levels of factual complexity.

\section{Mutual Nearest-Neighbor Metric}
\label{cha:metric}
For two models with representations $f$, $g$, the mutual $k$-nearest neighbor metric measures the average overlap of their respective nearest neighbor sets \citep{huh2024platonic}. According to the original definition, define $x_i \sim \mathcal{X}$ as a sample from the data distribution $\mathcal{X}$. $\{x_i\}_{i=1}^b$ is a mini-batch sampled from this data distribution. Two models $f$ and $g$ extract features $\phi_i = f(x_i)$ and $\psi_i = g(x_i)$. The collections of these features are denoted as $\Phi = \{\phi_1, \phi_2,...,\phi_b\}$ and $\Psi = \{\psi_1,\psi_2,...,\psi_b\}$. Then we compute the respective nearest neighbor sets $S(\phi_i)$ and $S(\psi_i)$ for each $x_i$ under the representations $f$ and $g$:
\begin{align}
    d_{knn}(\phi_i, \Phi \backslash \phi_i) = S(\phi_i); \\
    d_{knn}(\psi_i, \Psi \backslash \psi_i) = S(\psi_i),
\end{align}
where $d_{knn}$ returns the set of indices of its $k$-nearest neighbors. Then we measure its average intersection via 
\begin{equation}
    m_{\text{NN}}(\phi_i, \psi_i) = \frac{1}{k} \lvert S(\phi_i) \cap S(\psi_i) \rvert,
\end{equation}
where $\lvert \cdot \rvert$ denotes the size of the intersection. We use the hidden state of the last layer to represent the extracted feature of a prompt, and following the original paper \citep{huh2024platonic}, we set $k=10$ and $b=1,000$ (we take all data points if the total size of the data is less than $1,000$). We apply $l_2$ normalization to the features, then use the inner product kernel to measure the distance between two features. The alignment of two models is measured by $\frac{1}{b} \Sigma_{i=1}^b m_{\text{NN}}(\phi_i, \psi_i)$.